%% file: ICML_camera.tex
\documentclass[nohyperref]{article}

\usepackage{microtype}
\usepackage{graphicx}
\usepackage{subfigure}
\usepackage{booktabs} % for professional tables

\usepackage{hyperref}

\usepackage[accepted]{icml2022}

\usepackage{amsmath}
\usepackage{amssymb}
\usepackage{mathtools}
\usepackage{amsthm}

\usepackage[capitalize,noabbrev]{cleveref}

\usepackage[textsize=tiny]{todonotes}

\usepackage[utf8]{inputenc} % allow utf-8 input
\usepackage[T1]{fontenc}    % use 8-bit T1 fonts
\usepackage{url}            % simple URL typesetting
\usepackage{booktabs}       % professional-quality tables
\usepackage{amsfonts}       % blackboard math symbols
\usepackage{nicefrac}       % compact symbols for 1/2, etc.
\usepackage{microtype}      % microtypography

\usepackage{graphicx}
\usepackage{float}
\usepackage{sidecap}
\usepackage{adjustbox}
\usepackage{wrapfig}
\usepackage{tikz-cd}
\usepackage{amssymb}
\usepackage{amsmath}
\usepackage{amsthm}
\usepackage{soul}
\usepackage{tikz}
\usepackage{enumitem}
\usepackage{algorithm}
\usepackage{algorithmic}
\usepackage{caption}
\usepackage{siunitx}
\usepackage{makecell}
\usepackage{comment}
\usepackage{todonotes}
\usepackage{comment}
\usepackage{dcolumn}

\usepackage{pgf, tikz}
\usetikzlibrary{positioning}
\usetikzlibrary{arrows, automata}
\usetikzlibrary{shapes,snakes}

\usetikzlibrary{fit,positioning}

\newcommand\independent{\protect\mathpalette{\protect\independenT}{\perp}}
\def\independenT#1#2{\mathrel{\rlap{$#1#2$}\mkern2mu{#1#2}}}

\newcommand{\dependent}{ \not\!\perp\!\!\!\perp}

\makeatletter

\setbox0\hbox{$\xdef\scriptratio{\strip@pt\dimexpr
    \numexpr(\sf@size*65536)/\f@size sp}$}
    
\newcommand{\scriptveryshortarrow}[1][3pt]{{%
    \hbox{\rule[\scriptratio\dimexpr\fontdimen22\textfont2-.2pt\relax]
              {\scriptratio\dimexpr#1\relax}{\scriptratio\dimexpr.4pt\relax}}%
  \mkern-4mu\hbox{\let\f@size\sf@size\usefont{U}{lasy}{m}{n}\symbol{41}}}}

\newtheorem{definition}{Definition}
\newtheorem{proposition}{Proposition}

\icmltitlerunning{Action-Sufficient State Representation Learning for Control with Structural Constraints}

\begin{document}

\twocolumn[
\icmltitle{Action-Sufficient State Representation Learning for Control with Structural Constraints}



\icmlsetsymbol{equal}{*}

\begin{icmlauthorlist}
\icmlauthor{Biwei Huang}{equal,cmu}
\icmlauthor{Chaochao Lu}{equal,cambridge,mpi}
\icmlauthor{Liu Leqi}{cmu}
\icmlauthor{José Miguel Hernández-Lobato}{cambridge}
\icmlauthor{Clark Glymour}{cmu}
\icmlauthor{Bernhard Schölkopf}{mpi}
\icmlauthor{Kun Zhang}{ai,cmu}
\end{icmlauthorlist}

\icmlaffiliation{cmu}{Carnegie Mellon University}
\icmlaffiliation{cambridge}{University of Cambridge}
\icmlaffiliation{mpi}{Max Planck Institute for Intelligent Systems, Tübingen}
\icmlaffiliation{ai}{Mohamed bin Zayed University of Artificial Intelligence}

%\icmlcorrespondingauthor{Biwei Huang}{biweih@andrew.cmu.edu}
%\icmlcorrespondingauthor{Chaochao Lu}{cl641@cam.ac.uk}
\icmlcorrespondingauthor{Kun Zhang}{kunz1@cmu.edu}

\icmlkeywords{Machine Learning, ICML}

\vskip 0.3in
]



\printAffiliationsAndNotice{\icmlEqualContribution} % otherwise use the standard text.

\begin{abstract}
  Perceived signals in real-world scenarios are usually high-dimensional and noisy, and finding and using their representation that contains essential and sufficient information required by downstream decision-making tasks will help improve computational efficiency and generalization ability in the tasks. In this paper, we focus on partially observable environments and propose to learn a minimal set of state representations that capture sufficient information for decision-making, termed \textit{Action-Sufficient state Representations} (ASRs). We build a generative environment model for the structural relationships among variables in the system and present a principled way to characterize ASRs based on structural constraints and the goal of maximizing cumulative reward in policy learning. We then develop a structured sequential Variational Auto-Encoder to estimate the environment model and extract ASRs. Our empirical results on CarRacing and VizDoom demonstrate a clear advantage of learning and using ASRs for policy learning. Moreover, the estimated environment model and ASRs allow learning behaviors from imagined outcomes in the compact latent space to improve sample efficiency.
\end{abstract}

\section{Introduction}
% \vspace{-1.5mm}
State-of-the-art reinforcement learning (RL) algorithms leveraging deep neural networks are usually data hungry and lack interpretability. For example, to attain expert-level performance on tasks such as chess or Atari games, deep RL systems usually require many orders of magnitude more training data than human experts \citep{Atari_17}. One of the reasons is that our perceived signals in real-world scenarios, e.g., images, are usually high-dimensional and may contain much irrelevant information for decision-making of the task at hand. This makes it difficult and expensive for an agent to directly learn optimal policies from raw observational data. Fortunately, the underlying states that directly guide decision-making could be much lower-dimensional \citep{Causality_ML, consciousness_17}. One example is that when crossing the street, our decision on when to cross relies on the traffic lights. 
The useful state of traffic lights (e.g., its color) can be represented by a single binary variable, while the perceived image is high-dimensional. It is essential to extract and exploit such lower-dimensional states to improve the efficiency and interpretability of the decision-making process.  

Recently, representation learning algorithms have been designed to learn abstract features from high-dimensional and noisy observations. Exploiting the abstract representations, instead of the raw data, has been shown to perform subsequent decision-making more efficiently \citep{Representation_overview}. Representative methods along this line include deep Kalman filters \citep{DKF_15}, deep variational Bayes filters \citep{DVBF_16}, world models \citep{World_Model}, PlaNet \citep{PlaNet_18}, DeepMDP \citep{Gelada19_ICML}, stochastic latent actor-critic \citep{SLAC_19}, SimPLe \citep{SimPLe}, Bisimulation-based methods \citep{Amy_DBM20}, Dreamer \citep{DreamerV1, DreamerV2}, and others \citep{CURL_ICML20, pmlr-v119-shu20a}. Moreover, if we can properly model and estimate the underlying transition dynamics, then we can perform model-based RL or planning, which can effectively reduce interactions with the environment \citep{World_Model, PlaNet_18, DreamerV1, DreamerV2}.

Despite the effectiveness of the above approaches to learning abstract features, current approaches usually fail to take into account whether the extracted state representations are sufficient and necessary for downstream policy learning. State representations that contain insufficient information may lead to sub-optimal policies, while those with redundant information may require more samples and more complex models for training. 
We address this problem by modeling the generative process and selection procedure induced by reward maximization; by considering a generative environment model involving observed states, state-transition dynamics, and rewards, and explicitly characterizing structural relationships among variables in the RL system, we propose a principled approach to learning minimal sufficient state representations. We show that only the state dimensions that have direct or indirect edges to the reward variable are essential and should be considered for decision making. Furthermore, they can be learned by maximizing their ability to predict the action, given that the cumulative reward is included in the prediction model, while at the same time achieving their minimality w.r.t. the mutual information with observations as well as their dimensionality. 
The contributions of this paper are summarized as follows:
\begin{itemize}[leftmargin=*,align=left,itemsep=0pt,topsep=0pt]
 \item We construct a generative environment model, which includes the observation function, transition dynamics, and reward function, and explicitly characterizes structural relationships among variables in the RL system. 
 \item We characterize a minimal sufficient set of state representations, termed Action-Sufficient state Representations (ASRs), for the downstream policy learning by making use of structural constraints and the goal of maximizing cumulative reward in policy learning.
 \item In light of the characterization, we develop Structured Sequential Variational Auto-Encoder (SS-VAE), which explicitly encodes structural relationships among variables, for reliable identification of ASRs. %Note that in SS-VAE, the latent states are organized with structures, instead of being independent as in traditional VAEs.
 \item Accordingly, policy learning can be done separately from representation learning, and the policy function only relies on a set of low-dimensional state representations, which improve both model and sample efficiency. Moreover, the estimated environment model and ASRs allow learning behaviors from imagined outcomes in the compact latent space, which effectively reduce possibly risky explorations. %We show that model-based RL, based on the proposed environment model and the guarantees for sufficient and minimal state representations, achieves comparable performance compared to model-free RL. 
\end{itemize}

% \vspace{-1.5mm}
\section{Environment Model with Structural Constraints}
\label{sec:asr}
% \vspace{-2mm}
In order to characterize a set of minimal sufficient state representations for downstream policy learning, we first formulate a generative environment model in partially observable Markov decision process (POMDP), and then show how to explicitly embed structural constraints over variables in the RL system and leverage them.  

Suppose we have sequences of observations $\{\langle o_t, a_t, r_t  \rangle\}_{t=1}^T$, where $o_t \in \mathcal{O}$ denotes perceived signals at time $t$, such as high-dimensional images, with $\mathcal{O}$ being the observation space, $a_t \in \mathcal{A}$ is the performed action with $\mathcal{A}$ being the action space, and $r_t \in \mathcal{R}$ represents the reward variable with $\mathcal{R}$ being the reward space. We denote the underlying states, which are latent, by $\vec{s}_t \in \mathcal{S}$, with $\mathcal{S}$ being the state space. 
We describe the generating process of the environment model as follows:
\begin{equation}
\left \{
 \begin{array}{lll}
  o_t = f(\vec{s}_t, e_t), \\
  r_t = g(\vec{s}_{t-1}, a_{t-1}, \epsilon_t), \\
  \vec{s}_t = h(\vec{s}_{t-1}, a_{t-1}, \eta_t),
 \end{array}\right.
 \label{Eq: model1}
\end{equation}
where $f$, $g$, and $h$ represent the observation function, reward function, and transition dynamics, respectively, and $e_t$, $\epsilon_t$, and $\eta_t$ are corresponding independent and identically distributed (i.i.d.) random noises. The latent states $\vec{s}_t$ form an MDP: given $\vec{s}_{t-1}$ and $a_{t-1}$, $\vec{s}_t$ are independent of states and actions before $t-1$. Moreover, the action $a_{t-1}$ directly influences latent states $\vec{s}_t$, instead of perceived signals $o_t$, and the reward is determined by the latent states (and the action) as well. The perceived signals $o_t$ are generated from the underlying states $\vec{s}_t$, contaminated by random noise $e_t$. We also consider noise $\epsilon_t$ in the reward function to capture unobserved factors that may affect the reward, as well as measurement noise. 

It is commonplace that the action variable $a_{t-1}$ may not influence every dimension of $\vec{s}_t$, and the reward $r_t$ may not be influenced by every dimension of $\vec{s}_{t-1}$ as well, and furthermore there are structural relationships among different dimensions of $\vec{s}_t$. Figure \ref{Fig: generalmodel} gives an illustrative graphical representation, where $s_{3,t-1}$ influences $s_{2,t}$, $a_{t-1}$ does not have an edge to $s_{3,t}$, and among the states, only $s_{2,t-1}$ and $s_{3,t-1}$ have edges to $r_t$. We use $R_t =\sum_{\tau=t}^{\infty} \gamma^{\tau-t} r_{\tau}$ to denote the discounted cumulative reward starting from time $t$, where $\gamma \in [0,1]$ is the discounted factor that determines how much immediate rewards are favored over more distant rewards.

To reflect such constraints, we explicitly encode the graph structure over variables, including the structure over different dimensions of $\vec{s}$ and the structures from $a_{t-1}$ to $\vec{s}_t$, $\vec{s}_{t-1}$ to $r_t$, and $\vec{s}_t$ to $o_t$. Accordingly, we re-formulate \eqref{Eq: model1} as follows:
\begin{equation}
\left \{
\begin{array}{lll}
 o_t = f(D_{\vec{s} \scriptveryshortarrow o} \odot \vec{s}_t, e_t), \\
 r_t = g(D_{\vec{s} \scriptveryshortarrow r} \odot \vec{s}_{t-1}, D_{a \scriptveryshortarrow r} \odot a_{t-1}, \epsilon_t), \\
 s_{i,t} = h_i(D_{\vec{s}(\cdot,i)} \odot \vec{s}_{t-1}, D_{a \scriptveryshortarrow \vec{s}(\cdot,i)} \odot a_{t-1}, \eta_{i,t}),
\end{array} \right.
\label{Eq: model2}
\end{equation}
 for $i = 1,\cdots, d$, where $\vec{s}_t = (s_{1,t}, \cdots, s_{d,t})^{\top}$, $\odot$ denotes element-wise product, and $D_{(\cdot)}$ are binary matrices indicating the graph structure over variables. Specifically, $D_{\vec{s} \scriptveryshortarrow o} \in \{0,1\}^{d \times 1}$ represents the graph structure from $d$-dimensional $\vec{s}_t$ to $o_t$, $D_{\vec{s} \scriptveryshortarrow r}  \in \{0,1\}^{d \times 1}$ the structure from $\vec{s}_{t-1}$ to the reward variable $r_t$, $D_{a \scriptveryshortarrow r}  \in \{0,1\}$  the structure from the action variable $a_{t-1}$ to the reward variable $r_t$, $D_{\vec{s}}  \in \{0,1\}^{d \times d}$ denotes the graph structure from $d$-dimensional $\vec{s}_{t-1}$ to $d$-dimensional $\vec{s}_{t}$ and $D_{\vec{s}(\cdot,i)}$ is its $i$-th column, and $D_{a \scriptveryshortarrow \vec{s}}  \in \{0,1\}^{1 \times d}$ corresponds to the graph structure from $a_{t-1}$ to $\vec{s}_{t}$ with $D_{a \scriptveryshortarrow \vec{s}(\cdot,i)}$ representing its $i$-th column.  
For example, $D_{\vec{s}(j,i)}=0$ means that there is no edge from $s_{j,t-1}$ to $s_{i,t}$. Here, we assume that the environment model, as well as the structural constraints, is invariant across time instance $t$.

\input{fig2_new.tex}

%\paragraph{Remarks. } The formulated generative environment model in Eq. (\ref{Eq: model2}) is quite general, which also covers some special scenarios where the reward is only partially controllable by actions or the perceived signals contain only partial information of the ASRs.

% \vspace{-1.5mm}
\subsection{Minimal Sufficient State Representations}\label{Sec: Suff and Minimal}
% \vspace{-1.5mm}
% summary of this section: two ways to learn minimal and sufficient state representations
Given observational sequences $\{\langle o_t, a_t, r_t \rangle\}_{t=1}^T$, we aim to learn minimal sufficient state representations for the downstream policy learning. In the following, we first characterize the state dimensions that are indispensable for policy learning, when the environment model, including structural relationships, is given. Then we provide criteria to achieve sufficiency and minimality of the estimated state representations, when only $\{\langle o_t, a_t, r_t \rangle\}_{t=1}^T$, but not the environment model, is given. 
% identify a minimal set of state dimensions that suffice for policy learning, when the environment model is given

\paragraph{Finding minimal sufficient state dimensions with a given environment model.}
RL agents learn to choose appropriate actions according to the current state vector $\vec{s}_t$ to maximize the future cumulative reward, in which some dimensions may be redundant for policy learning.
Then how can we identify a minimal subset of state dimensions that are sufficient to choose optimal actions? Below, we first give the definition of \textit{Action-Sufficient state Representations (ASRs)} according to the graph structure. We further show in Proposition \ref{Proposition: ASR} that ASRs are minimal sufficient for policy learning, and they can be characterized by leveraging the (conditional) independence/dependence relations among the quantities, under the Markov condition and faithfulness assumption \citep{Pearl00, SGS93}. 
% \begin{definition}[Action-Sufficient State Representations (ASRs)]
%   Given the graphical representation corresponding to the environment model, such as the representation in Figure \ref{Fig: generalmodel}, we define ASRs $\vec{s}_t^{\,\text{ASR}} \subseteq \vec{s}_t$ as those dimensions which have a directed path\footnote{A directed path in a directed graph is a sequence of edges which joins a sequence of distinct vertices, and the edges are all directed in the same direction.}, that does not go through any action variable, to $r_{t+\tau}$, for $\tau>0$, .
%     \label{define: ASR}
% \end{definition}
\begin{definition}[Action-Sufficient State Representations (ASRs)]
  Given the graphical representation corresponding to the environment model, such as the representation in Figure \ref{Fig: generalmodel}, we define recursively ASRs that affect the future reward as: (1) $s_{i,t} \in \vec{s}_t^{\,\text{ASR}}$ has an edge to the reward in the next time-step $r_{t+1}$, or (2) $s_{i,t} \in \vec{s}_t^{\,\text{ASR}}$ has an edge to another state dimension in the next time-step $s_{j,t+1}$, such that the same component at time $t$ is in ASRs, i.e., $s_{j,t} \in \vec{s}_t^{\,\text{ASR}}$.
    \label{define: ASR}
\end{definition}

\begin{proposition}
Under the assumption that the graphical representation, corresponding to the environment model, is Markov and faithful to the measured data, $\vec{s}_t^{\,\text{ASR}} \subseteq \vec{s}_t$ are a minimal subset of state dimensions that are sufficient for policy learning, and $s_{i,t} \in \vec{s}_t^{\,\text{ASR}}$ if and only if $s_{i,t} \dependent R_{t+1} | a_{t-1:t}, \vec{s}_{t-1}^{\,\text{ASR}}$.
  \label{Proposition: ASR}
\end{proposition}

%In the above proposition, sufficiency is guaranteed by involving all dimensions that are conditionally dependent on the action variable in ASRs, and the minimal set is guaranteed by excluding any dimension that is conditionally independent of the action variable from ASRs. 
% \vspace{-2mm}
Proposition \ref{Proposition: ASR} can be shown based on the global Markov condition and the faithfulness assumption, which connects d-separation\footnote{A path $p$ is said to be d-separated by a set of nodes $Z$ if and only if (1) $p$ contains a chain $i \rightarrow m \rightarrow j$ or a fork $i \leftarrow m \rightarrow j$ such that the middle node $m$ is in $Z$, or (2) $p$ contains a collider $i \rightarrow m \leftarrow j$ such that the middle node $m$ is not in $Z$ and such that no descendant of $m$ is in $Z$.}  to conditional independence/dependence relations. A proof is given in Appendix. According to the above proposition, it is easy to see that for the graph given in Figure \ref{Fig: generalmodel}, we have $\vec{s}_t^{\,\text{ASR}} = (s_{2,t}, s_{3,t})^{\top}$. That is, we only need $(s_{2,t}, s_{3,t})^{\top}$, instead of $\vec{s}_t$, for the downstream policy learning.

% minimal sufficient state representations based on two conditions: action prediction + structural constraints (I am not sure whether this part is clear)
% \vspace{-2.5mm}
\paragraph{Minimal sufficient state representation learning from observed sequences. }In practice, we usually do not have access to the latent states or the environment model, but instead only the observed sequences $\{\langle o_t, a_t, r_t  \rangle\}_{t=1}^T$. Then how can we learn the ASRs from the raw high-dimensional inputs such as images? We denote by $\tilde{\vec{s}}_t$ the estimated whole latent state representations and $\tilde{\vec{s}}_t^{\,\text{ASR}} \subseteq \tilde{\vec{s}}_t$ the estimated minimal sufficient state representations for policy learning. 

As discussed above, ASRs and $R_{t+1}$ are dependent given $a_{t-1:t}$ and $\tilde{\vec{s}}_{t-1}^{\,\text{ASR}}$, while other state dimensions are independent of $R_{t+1}$, so we can learn the ASRs by maximizing\footnote{Note that, when observational sequences are generated from a random policy (i.e., $\vec{s}_t \independent a_t$), learning the ASRs can be also implemented in a simpler way, which is given in Appendix. }
\begin{equation}
\resizebox{.85\hsize}{!}{$I(\tilde{\vec{s}}^{\,\text{ASR}}; R_{t+1} | a_{t-1:t}, \tilde{\vec{s}}_{t-1}^{\,\text{ASR}}) - I(\tilde{\vec{s}}^{\,\text{C}}; R_{t+1} | a_{t-1:t}, \tilde{\vec{s}}_{t-1}^{\,\text{ASR}})$},
\end{equation}
where $\tilde{\vec{s}}^{\,\text{C}} = \tilde{\vec{s}} \backslash \tilde{\vec{s}}^{\,\text{ASR}}$, and $I(\cdot)$ denotes mutual information. Such regularization is used to achieve minimal sufficient state representations for policy learning; that is, only $\tilde{\vec{s}}_{t-1}^{\,\text{ASR}}$ are useful for policy learning, while $\tilde{\vec{s}}_{t-1}^{\,\text{C}}$ are not. Furthermore, the mutual information can be represented as a form of conditional entropy, e.g.,  $I(\tilde{\vec{s}}^{\,\text{ASR}}; R_{t+1} | a_{t-1:t}, \vec{s}_{t-1}^{\,\text{ASR}}) = H(\tilde{\vec{s}}^{\,\text{ASR}}|a_{t-1:t}, \vec{s}_{t-1}^{\,\text{ASR}}) - H(\tilde{\vec{s}}^{\,\text{ASR}}|R_{t+1}, a_{t-1:t}, \vec{s}_{t-1}^{\,\text{ASR}})$, where $H(\cdot)$ denotes the conditional entropy, with
\begin{equation*}
\begin{array}{lll}
\footnotesize
     & H(\tilde{\vec{s}}^{\,\text{ASR}}|a_{t-1:t}, \tilde{\vec{s}}_{t-1}^{\,\text{ASR}})\\
     &=\! - \mathbb{E}_{q_{\phi}} \mathbb{E}_{p_{\alpha_1}} \! \big\{ \!\log p_{\alpha_1}( \tilde{\vec{s}}^{\text{ASR}} |a_{t-1:t},\tilde{\vec{s}}_{t-1}^{\,\text{ASR}})\! \big \} \\
     &=\! - \mathbb{E}_{q_{\phi}} \mathbb{E}_{p_{\alpha_1}} \! \big\{ \!\log p_{\alpha_1}( \tilde{\vec{s}}^{\text{ASR}} |a_{t-1:t},\tilde{D}^{ASR} \odot \tilde{\vec{s}}_{t-1})\! \big \}\!,
\end{array}
\end{equation*}
and
\begin{equation*}
\begin{array}{lll}
\footnotesize
& H(\tilde{\vec{s}}^{\,\text{ASR}}|R_{t+1},a_{t-1:t}, \tilde{\vec{s}}_{t-1}^{\,\text{ASR}}) \\
&\!\!\!\!\!=\! - \mathbb{E}_{q_{\phi}} \mathbb{E}_{p_{\alpha_2}} \! \big\{ \!\log p_{\alpha_2}(\tilde{\vec{s}}^{\text{ASR}} |R_{t+1},a_{t-1:t},\tilde{D}^{ASR} \odot \tilde{\vec{s}}_{t-1})\! \big \}, \\
\end{array}
\end{equation*}

where $p_{\alpha_i}$, for $i=1,2$, denotes the probabilistic predictive model of $\tilde{\vec{s}}^{\,\text{ASR}}$ with parameters $\alpha_i$, and $q_{\phi}(\tilde{\vec{s}}_t | \tilde{\vec{s}}_{t-1}, \mathbf{y}_{1:t},a_{1:t-1})$ is the probabilistic inference model of $\tilde{\vec{s}}_t$ with parameters $\phi$ and $\mathbf{y}_t \!= \!(o_t^T, r_t^T)$, and $\tilde{D}^{\,\text{ASR}} \in \{0,1\}^{\tilde{d} \times 1}$ is a binary vector indicating which dimensions of $\tilde{\vec{s}}_t$ are in $\tilde{\vec{s}}_t^{\,\text{ASR}}$, so $\tilde{D}^{\,\text{ASR}} \odot \tilde{\vec{s}}_t$ gives ASRs $\tilde{\vec{s}}_t^{\,\text{ASR}}$. Similarly, we can also represent $I(\tilde{\vec{s}}^{\,\text{C}}; R_{t+1} | a_{t-1:t}, \tilde{\vec{s}}_{t-1}^{\,\text{ASR}})$ in such a way.

Although with the above regularization, we can achieve ASRs theoretically, in practice, we further add another regularization to achieve minimality of the representation by minimizing conditional mutual information between observed high-dimensional signals $\mathbf{y}_t$ and the ASR $\tilde{\vec{s}}_t^{\,\text{ASR}}$ at time $t$ given data at previous time instances, similar to that in information bottleneck \citep{IB_99}, and meanwhile minimizing the dimensionality of ASRs with sparsity constraints:
 \begin{equation*}
     \lambda_1 \sum\nolimits_{t=2}^T I(\mathbf{y}_t; \tilde{\vec{s}}_t^{\,\text{ASR}} | \mathbf{y}_{1:t-1}, a_{1:t-1}, \tilde{\vec{s}}_{t-1}) + \lambda_2 \|\tilde{D}^{\,\text{ASR}}\|_1,
 \end{equation*}
where the conditional mutual information can be upper bound by a KL-divergence:
\begin{align}
I(\mathbf{y}_t; \tilde{\vec{s}}_t^{\,\text{ASR}} | \mathbf{y}_{1:t-1}, a_{1:t-1}, \tilde{\vec{s}}_{t-1}) 
      \leq  \mathbb{E}\big\{\text{KL} \big( q_{\phi'} \Vert p_{\gamma}\big)\big\},
\end{align}
with $q_{\phi'} \equiv q_{\phi'}(\tilde{\vec{s}}_t^{\,\text{ASR}} | \tilde{\vec{s}}_{t-1}, \mathbf{y}_{1:t},a_{1:t-1})$ and  $p_{\gamma} \equiv p_{\gamma}(\tilde{\vec{s}}_t^{\,\text{ASR}}|\tilde{\vec{s}}_{t-1},a_{t-1}; D_{\vec{s}},D_{a \scriptveryshortarrow \vec{s}})$ being the transition dynamics of $\tilde{\vec{s}}_t$ with parameters $\gamma$, and the expectation is over $p(\tilde{\vec{s}}_{t-1}, \mathbf{y}_{1:t},a_{1:t-1})$.

Furthermore, Proposition \ref{Proposition: ASR} shows that given the (estimated) environment model, only those state dimensions that have a directed path to the reward variable are the ASRs. In our learning procedure, we also take into account the relationship between the learned states $\tilde{\vec{s}}_t$ and the reward, and leverage such structural constraints for learning the ASRs. Denote by $\check{D}^{ASR} \in \{0,1\}^{\tilde{d} \times 1}$ a binary vector indicating whether the corresponding state dimension in $\tilde{\vec{s}}_t$ has a directed path to the reward variable. 
Consequently, we enforce the similarity between $\check{D}^{ASR}$ and $\tilde{D}^{ASR}$ by adding an $L_1$ norm on $\check{D}^{ASR}- \tilde{D}^{ASR}$. 
Therefore, the ASRs can be learned by maximizing the following function: 
\begin{equation}\small
\begin{array}{lll}
    \mathcal{L}^{\text{min \& suff}}=\\
      \!\!\!\! \lambda_3 \! \underbrace{\sum \nolimits_{t=1}^{T} \!\!\big\{      I(\tilde{\vec{s}}^{\,\text{ASR}}\!; R_{t+1} | a_{t-1:t}, \tilde{\vec{s}}_{t-1}^{\,\text{ASR}}) \!-\! I(\tilde{\vec{s}}^{\,\text{C}}\!; R_{t+1} | a_{t-1:t}, \tilde{\vec{s}}_{t-1}^{\,\text{ASR}})\! \big\}}_{\text{Sufficiency \& Minimality}} \\
      \!\!\!\!- \lambda_4 \|\check{D}_{\,\text{ASR}}\! -\! \tilde{D}_{\,\text{ASR}}\|_1 
     \!\!\underbrace{-\! \lambda_1 \sum_{t=1}^T \mathbb{E} \big\{\text{KL} \big( q_{\phi} \Vert p_{\gamma} \big) \big\} \!-\! \lambda_2 \|\tilde{D}^{ASR}\|_1}_{\text{Further restrictions of minimality}},
\end{array}
\label{Suff&Min}
% \vspace{-2mm}
\end{equation}
where $\lambda$'s are regularization terms, and note that $\check{D}^{ASR}$ can be directly derived from the estimated structural matrices $D_{a \scriptveryshortarrow r} \text{ and } D_{\vec{s}(\cdot,i)}$. The constraint in Eq. \ref{Suff&Min} provides a principled way to achieve minimal sufficient state representations. Notice that it is just part of the objective function to maximize, and it will be involved in the complete objective function in Fig.~\ref{Obj} to learn the whole environment model.

% \vspace{-3mm}
\paragraph{Remarks.} 
By explicitly involving structural constraints, we achieve minimal sufficient state representations from the view of generative process underlying the RL problem and the selection procedure induced by reward maximization, which enjoys the following advantages. 1) The structural information provides an interpretable and intuitive picture of the generating process. 2) Accordingly, it also provides an interpretable and intuitive way to characterize a minimal sufficient set of state representations for policy learning, which removes unrelated information. 3) There is no information loss when representation learning and policy learning are done separately, which is computationally more efficient. 4) The generative environment model is fixed, independent of the behavior policy that is performed. Furthermore, based on the estimated environment model and ASRs, it is flexible to use a wide range of policy learning methods, and one can also perform model-based RL, which effectively reduce possibly risky explorations.  

% \vspace{-1mm}
\section{Structured Sequential VAE for the Estimation of ASRs}\label{Sec: estimation}
% \vspace{-.5mm}
 In this section, we give estimation procedures for the environment model and ASRs, as well as the identifiability guarantee in linear cases. 

% \vspace{-2.5mm}
% identifiability in linear case
\paragraph{Identifiability in Linear-Gaussian Cases.} 
Below, we first show the identifiability guarantee in the linear case, as a special case of Eq.  \eqref{Eq: model2}:
\begin{equation}
\left \{
\begin{array}{lll}
o_t = D_{\vec{s} \scriptveryshortarrow o}^{\top} \vec{s}_t +  e_t, \\
r_{t+1} = D_{\vec{s} \scriptveryshortarrow r}^{\top} \vec{s}_{t} + D_{a \scriptveryshortarrow r}^{\top} a_{t} + \epsilon_{t+1}, \\
\vec{s}_t = D_{\vec{s}}^{\top} \vec{s}_{t-1} + D_{a \scriptveryshortarrow \vec{s}}^{\top} a_{t-1} + \eta_t.
\end{array} \right.
\label{Eq: linear}
\end{equation}
In the linear case, $D_{\vec{s} \scriptveryshortarrow o}$, $D_{\vec{s} \scriptveryshortarrow r}$, $D_{a \scriptveryshortarrow r}$, $D_{\vec{s}}$, and $D_{a  \scriptveryshortarrow\vec{s}}$ are linear coefficients, indicating corresponding graph structures and also the strength. Denote the covariance matrices of $e_t$ and $\epsilon_t$ by $\Sigma_{e}$ and $\Sigma_{\epsilon}$, respectively. 
Further let $\ddot{D}_{\vec{s} \scriptveryshortarrow o} :=(D_{\vec{s} \scriptveryshortarrow o}^{\top},D_{\vec{s} \scriptveryshortarrow r}^{\top})^{\top}$. The following proposition shows that the environment model in the linear case is identifiable up to some orthogonal transformation on certain coefficient matrices from  observed data $\{\langle o_t,a_t,r_t  \rangle\}_{t=1}^T$.

% \vspace{1mm}
\begin{proposition}[Identifiability]
  Suppose the perceived signal $o_t$, the reward $r_t$, and the latent states $\vec{s}_t$ follow a linear environment model. If assumptions A1$\sim$A4 (given in Appendix \ref{sec:assumption-thm-1}) hold and with the second-order statistics of the observed data $\{\langle o_t,a_t,r_t  \rangle\}_{t=1}^T$, the noise variances $\Sigma_e$ and $\Sigma_{\epsilon}$, $D_{a \scriptveryshortarrow r}$, $\ddot{D}_{\vec{s} \scriptveryshortarrow o}^{\top}{D_{\vec{s}}}^k {D_{a \scriptveryshortarrow \vec{s}}}^{\top}$ (with $k \geq 0$), and $\ddot{D}_{\vec{s} \scriptveryshortarrow o}^{\top} \ddot{D}_{\vec{s} \scriptveryshortarrow o}$ are uniquely identified. 
  \label{Pro: identifiability}
\end{proposition}

% \vspace{-2mm}
% SSVAE in general nonlinear cases
This proposition shows that in the linear case, with the second-order statistics of the observed data, we can identify the parameters up to orthogonal transformations. In particular, suppose the linear environment model with parameters $({D_{\vec{s} \scriptveryshortarrow o}},{D_{\vec{s} \scriptveryshortarrow r}},{D_{a \scriptveryshortarrow r}}, {D_{\vec{s}}},{D_{a \scriptveryshortarrow \vec{s}}},\Sigma_e,\Sigma_{\epsilon})$ and that with $(\tilde{D}_{\vec{s} \scriptveryshortarrow o},\tilde{D}_{\vec{s} \scriptveryshortarrow r},\tilde{D}_{a \scriptveryshortarrow r},\tilde{D}_{\vec{s}},\tilde{D}_{a \scriptveryshortarrow \vec{s}},\tilde{\Sigma}_{\tilde{e}},\tilde{\Sigma}_{\tilde{\epsilon}})$ are observationally equivalent. Then we have $\tilde{\ddot{D}}_{\vec{s} \scriptveryshortarrow o} \!=\!U  \ddot{D}_{\vec{s} \scriptveryshortarrow o}$, $\tilde{D}_{a \scriptveryshortarrow r}\!\! =\!\! D_{a \scriptveryshortarrow r}$, $\tilde{D}_{\vec{s}}\! = \!U^{\top} {D_{\vec{s}}} U$, $\tilde{D}_{a \scriptveryshortarrow \vec{s}} \!=\!{D_{a \scriptveryshortarrow \vec{s}}} U$, $\tilde{\Sigma}_{\tilde{e}} \!=\! \Sigma_e$, and $\tilde{\Sigma}_{\tilde{\epsilon}} \!=\! \Sigma_{\epsilon}$, where $U$ is an orthogonal matrix. 

\begin{figure*}[t] 
% % \vspace{-2mm}
\centering
\begin{equation*} 
\small
  \begin{array}{lll}
    & \mathcal{L}(\mathbf{y}_{1:T}; (\theta, \phi, \gamma, \alpha, D_{(\cdot)}))\\
    &= \sum \nolimits_{t=1}^{T-2} \mathbb{E}_{q_{\phi}} \big\{ \underbrace{\log p_{\theta}( o_t |\tilde{\vec{s}}_t; D_{\vec{s} \scriptveryshortarrow o}) + \log p_{\theta}(r_{t+1} |\tilde{\vec{s}}_t, a_{t}; D_{\vec{s} \scriptveryshortarrow r}, D_{a \scriptveryshortarrow r})}_{\text{Reconstruction}} 
     + \underbrace{\log p_{\theta}(o_{t+1}|
    \tilde{\vec{s}}_{t}) + \log p_{\theta}(r_{t+2}|\tilde{\vec{s}}_{t}, a_{t+1})}_{\text{Prediction}} \big \} \\
    %  & + \lambda_3 \underbrace{\sum \nolimits_{t=1}^{T} \big\{ \mathbb{E}_{q_{\phi\!,\alpha_1}}\! \big\{\!\log p_{\alpha_1}( a_t |\tilde{D}^{ASR} \odot \tilde{\vec{s}}_t\!, R_{t+1},\tilde{\vec{s}}_{t-1}) \!\big \}\mathbb{E}_{q_{\phi\!,\alpha_2}} \big\{\log p_{\alpha_2}( a_t | R_{t+1},\tilde{\vec{s}}_{t-1}) \big \}  \big\}}_{\text{Sufficiency}}\\
     &~~~~~~~~~~ +\lambda_3 \underbrace{\sum \nolimits_{t=1}^{T} \!\big\{      I(\tilde{\vec{s}}^{\,\text{ASR}}\!; R_{t+1} | a_{t-1:t}, \tilde{\vec{s}}_{t-1}^{\,\text{ASR}}) \!-\! I(\tilde{\vec{s}}^{\,\text{C}}\!; R_{t+1} | a_{t-1:t}, \tilde{\vec{s}}_{t-1}^{\,\text{ASR}})\! \big\}}_{\text{Sufficiency \& Minimality}} \\
    & ~~~~~~~~~~~ - \lambda_1 \sum \nolimits_{t=1}^{T} \underbrace{\mathbb{E}\big\{\text{KL} \big( q_{\phi'}(\tilde{\vec{s}}_t^{\text{ASR}} | \tilde{\vec{s}}_{t-1}, \mathbf{y}_{1:t},a_{1:t-1}) \Vert \underbrace{p_{\gamma}(\tilde{\vec{s}}_t^{\text{ASR}}|\tilde{\vec{s}}_{t-1},a_{t-1}; D_{\vec{s}},D_{a \scriptveryshortarrow \vec{s}})}_{\text{Transition}} \big)\big\} - \lambda_2 \| \tilde{D}_{\,\text{ASR}}\|_1}_{\text{Conditional disentanglement \& Further restrictions of minimality}} \\
    & ~~~~~~~~~~~ - \underbrace{\big(\lambda_5 \| {D_{\vec{s} \scriptveryshortarrow o}} \|_1 + \lambda_6 \| {D_{\vec{s} \scriptveryshortarrow r}} \|_1 + \lambda_7 \| {D_{\vec{s}}} \|_1 + \lambda_8 \| {D_{a \scriptveryshortarrow \vec{s}}} \|_1  + \lambda_4 \|\check{D}_{\,\text{ASR}} - \tilde{D}_{\,\text{ASR}}\|_1 \big)}_{\text{Sparsity}} ,
  \end{array}
\end{equation*}
% \vspace{5mm}
\caption{Our objective function.}\label{Obj} \vspace{-2mm}
\end{figure*}

% \vspace{-2mm}
\paragraph{General Nonlinear Cases.} %The identifiability guarantee in general nonlinear cases is nontrivial, so we focus on its practical implementation. 
To handle general nonlinear cases with the generative process given in Eq.~\eqref{Eq: model2}, we develop a Structured Sequential VAE (SS-VAE) to learn the model (including structural constraints) and infer latent state representations $\tilde{\vec{s}}_t$ and ASRs $\tilde{\vec{s}}_t^{\,\text{ASR}}$, with the input $\{\langle o_t,a_t,r_t  \rangle\}_{t=1}^T$. Specifically, the latent state dimensions are organized with structures, captured by $D_{\vec{s}}$, to achieve conditional independence. The structural relationships over perceived signals, latent states, the action variable, and the reward variable are also embedded as free parameters (i.e., $D_{\vec{s} \scriptveryshortarrow o}, D_{\vec{s} \scriptveryshortarrow r}, D_{a \scriptveryshortarrow r}, D_{a \scriptveryshortarrow \vec{s}}$) into SS-VAE. Moreover, we aim to learn state representations $\tilde{\vec{s}}_t$ and ASRs $\tilde{\vec{s}}_t^{\,\text{ASR}}$ that satisfy the following properties: (\textit{i}) $\tilde{\vec{s}}_t$ should capture sufficient information of observations $o_{t}$, $r_{t}$, and $a_t$, that is, it should be enough to enable reconstruction. (\textit{ii}) The state representations should allow for accurate predictions of the next state and also the next observation. (\textit{iii}) The transition dynamics should follow an MDP. (\textit{iv}) $\tilde{\vec{s}}_t^{\,\text{ASR}}$ are minimal sufficient state representations for the downstream policy learning.

Let $\mathbf{y}_{1:T} = \{ (o_t^{\top},r_t^{\top})^{\top} \}_{t=1}^T$. To achieve the above properties, we maximize the objective function shown in Fig.~\ref{Obj}, %which contains a lower bound of $\log P(\mathbf{y}_{1:T})$, prediction loss, and constraints to achieve minimal sufficient state representations and an MDP over state representations:
which contains the reconstruction error at each time instance, the one-step prediction error of observations, the KL divergence to constrain the latent space, and moreover, the MDP restrictions on transition dynamics, the sufficiency and minimality guarantee of state representations for policy learning, as well as sparsity constraints on the graph structure. 
We denote by $p_{\theta}$ the generative model with parameters $\theta$ and structural constraints $D_{(\cdot)}$, $q_{\phi}$ the inference model with parameters $\phi$, $p_{\gamma}$ the transition dynamics, and $p_{\alpha_i}$ the predictive model of ASRs. Each factor in $p_{\gamma}$, $q_{\phi}$, and $p_{\alpha_i}$ is modeled with a mixture of Gaussians (MoGs), to approximate a wide class of continuous distributions. 

% \begin{SCfigure*}[0.8][t]
\begin{figure*}
% \vspace{-2mm}
    \centering
    \includegraphics[width=0.6\textwidth]{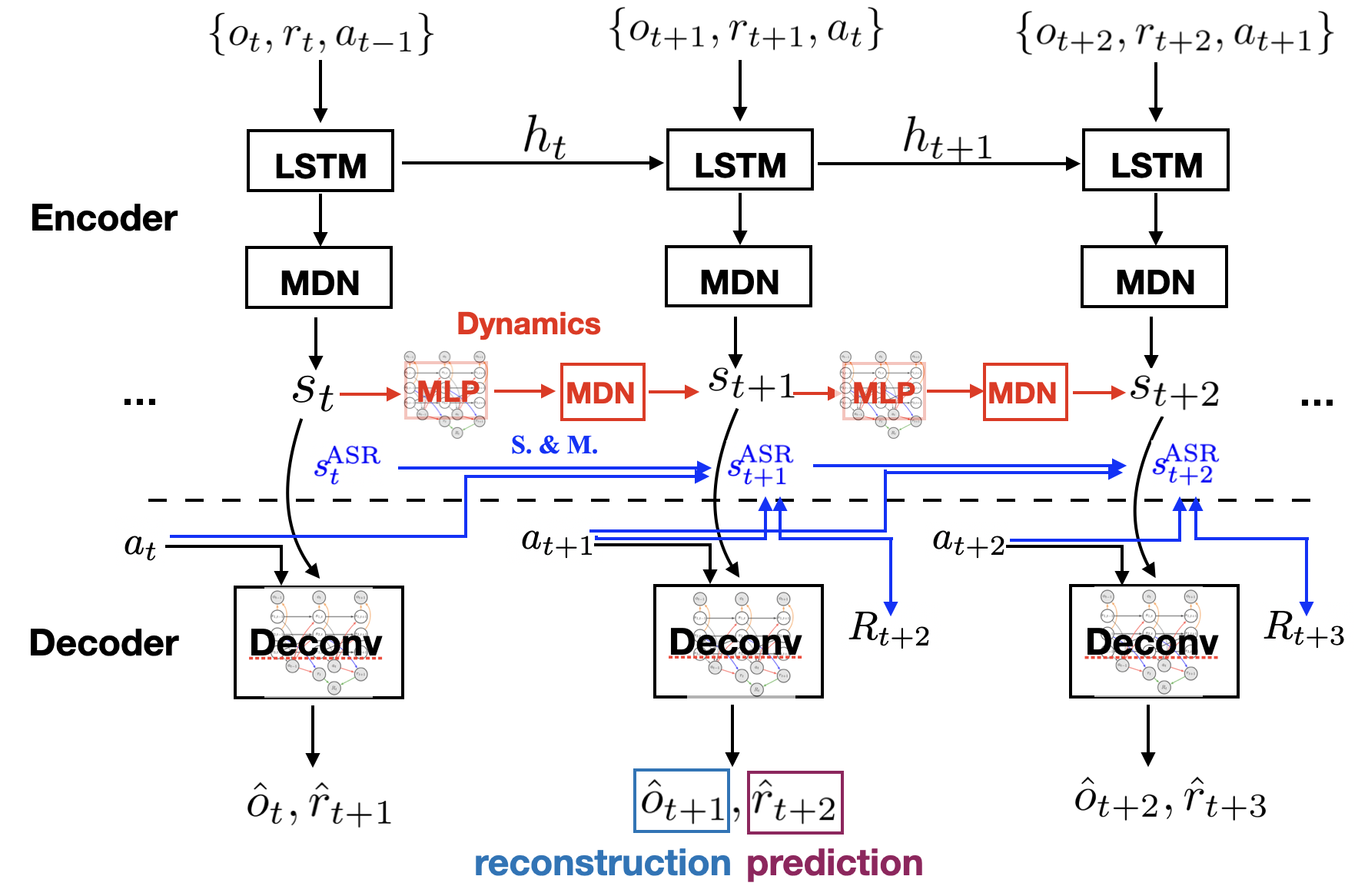}
    \caption{Diagram of neural network architecture to learn state representations. The corresponding structural constraints are involved in ``Deconv'' and ``MLP'', and ``S.\&M." represents the regularization part for minimal sufficient state representation learning. }
    \label{Fig: architecture-diagram}
    % \vspace{-1mm}
\end{figure*}
% \end{SCfigure*}

Below are the details of each component in the above objective function:
\begin{itemize}[leftmargin=*,align=left,itemsep=1.2pt,topsep=0pt]
    \item Reconstruction and prediction components: These two parts are commonly used in sequential VAE. They aim to minimize the reconstruction error and prediction error of the perceived signal $o_t$ and the reward $r_t$. 
    \item Transition component: To achieve the property that state representations satisfy an MDP, we explicitly model the transition dynamics: $\log p_{\gamma}(\tilde{\vec{s}}_t | \tilde{\vec{s}}_{t-1}, a_{t-1};D_{\vec{s}},D_{a \scriptveryshortarrow \vec{s}})$. In particular,  $\tilde{\vec{s}}_t| \tilde{\vec{s}}_{t-1}$ is modelled with a mixture of Gaussians: $\sum_{k=1}^K \! \pi_k \mathcal{N}\big(\boldsymbol{\mu}_k(\tilde{\vec{s}}_{t-1},\! a_{t-1}),\Sigma_k(\tilde{\vec{s}}_{t-1},\! a_{t-1})\big)$, where $K$ is the number of mixtures, $\boldsymbol{\mu}_k(\cdot)$ and $\Sigma_k(\cdot)$ are given by multi-layer perceptrons (MLP) with inputs $\tilde{\vec{s}}_{t-1}$ and $a_{t-1}$, parameters $\gamma$, and structural constraints ${D_{\vec{s}}}$ and ${D_{a \scriptveryshortarrow \vec{s}}}$. This explicit constraint on state dynamics is essential for establishing a Markov chain in latent space and for learning a representation for long-term predictions. Note that unlike in traditional VAE \citep{VAE_13}, we do not assume that different dimensions in $\tilde{\vec{s}}_t$ are marginally independent, but model their structural relationships explicitly to achieve conditional independence. 
    \item KL-divergence constraint: The KL divergence is used to constrain the state space with multiple purposes: (1) It is used in the lower bound of $\log P(\mathbf{y}_{1:T})$ to achieve conditional disentanglement between $q_{\phi}(\tilde{s}_{i,t}|\cdot)$ and $q_{\phi}(\tilde{s}_{j,t}|\cdot)$ for $i \neq j$, (2) and also to achieve further restrictions of minimality of ASRs.
    \item Sufficiency \& minimality constraints: We achieve minimal sufficient state representations for the downstream policy learning by leveraging the conditional mutual information between $\tilde{\vec{s}}_t^{\,\text{ASR}}$ and $R_{t+1}$, and structural constraints. For details, please refer to Section \ref{Sec: Suff and Minimal}.  
    \item Sparsity constraints: According to the edge-minimality property \citep{Minimality_Zhang}, we additionally put sparsity constraints on structural matrices to achieve better identifiability. In particular, we use $L_1$ norm of the structural matrices as regularizers in the objective function to achieve sparsity of the solution.
\end{itemize}

Figure~\ref{Fig: architecture-diagram} gives the diagram of the neural network architecture in model training. 
We use SS-VAE to learn the environment model and ASRs. Specifically, the encoder, which is used to learn the inference model $q_{\phi}(\tilde{\vec{s}}_t | \tilde{\vec{s}}_{t-1}, \mathbf{y}_{1:t},a_{1:t-1})$, includes a Long Short-Term Memory (LSTM \citep{LSTM_97}) to encode the sequential information with output $h_t$ and a Mixture Density Network (MDN \citep{MDN_Bishop}) to output the parameters of  MoGs. At each time instance, the input $\langle o_{t+1},r_{t+1},a_t \rangle$ is projected to the encoder and a sample of $\tilde{\vec{s}}_{t+1}$ is inferred from $q_{\phi}$ as output. The generated sample further acts as an input to the decoder, together with $a_{t+1}$ and structural matrices $D_{\vec{s} \scriptveryshortarrow o}, D_{\vec{s} \scriptveryshortarrow r}$, and $D_{a \scriptveryshortarrow r}$. Then the decoder outputs $\hat{o}_{t+1}$ and $\hat{r}_{t+2}$.
Moreover, the state dynamics which satisfies a Markov process and is embedded with structural constraints $D_{\vec{s}}$ and $D_{a \scriptveryshortarrow \vec{s}}$, is modeled with an MLP and MDN, marked with red in Figure~\ref{Fig: architecture-diagram}. The part for minimal sufficient representations (denoted by \textit{S.\&M.}) uses MLP and is marked with blue. During training, we approximate the expectation in $\mathcal{L}$ by sampling and then jointly learn all parameters by maximizing $\mathcal{L}$ using stochastic gradient descent. %Please refer to Appendix for more details.

% \begin{figure}[t]
%     \centering
%     	\setlength{\abovecaptionskip}{-1pt}
% 	\setlength{\belowcaptionskip}{-1pt}
%     \includegraphics[width=0.6\textwidth]{figs/diagram_new2.png}
%     \caption{Diagram of neural network architecture to learn state representations. The corresponding structural constraints are involved in ``Deconv'' and ``MLP'', and ``AP" represents the action prediction part for sufficient state representation learning. }
%     \label{Fig: architecture-diagram}
%     \vspace{-3mm}
% \end{figure}
% \begin{figure}[htp!]
%     \centering
% %     	\setlength{\abovecaptionskip}{0pt}
% % 	\setlength{\belowcaptionskip}{0pt}
%     \includegraphics[width=0.7\textwidth]{figs/diagram_RL.png}
%     \caption{Diagram of neural network architecture to learn state representations. The corresponding structural constraints are involved in ``Deconv'' and ``MLP''. }
%     \label{Fig: architecture-diagram}
% \end{figure}

%\paragraph{Inference}
% \vspace{-1mm}
\section{Policy Learning with ASRs}
% \vspace{-.5mm}

After estimating the generative environment model, we are ready to learn the optimal policy, where the policy function only depends on low-dimensional ASRs, instead of high-dimensional images. The entire procedure roughly contains the following three parts: (1) data collection with a random or sub-optimal policy, (2) environment model estimation (with details in Section \ref{Sec: estimation}), and (3) policy learning with ASRs. 
%Moreover, in complex environments, we may need to iteratively collect new experience and refine the environment model, that is, to repeat the three steps until convergence. 
Notably, the generative environment model is fixed, regardless of the behavior policy that is used to generate the data, and after learning the environment model, as well as the inference model for ASRs, our framework is flexible for both model-free and model-based policy learning.

% \vspace{-.5mm}
\paragraph{Model-Free Policy Learning. }
For model-free policy learning, we make use of the learned environment model to infer ASRs $\tilde{\vec{s}}_t^{\,\text{ASR}}$ from past observed sequences $\{o_{\leq t},r_{\leq t},a_{\leq t-1} \}$ and then predict the action with the estimated low-dimensional ASRs. 
Our method is flexible to use a wide range of model-free methods; for example, one may use deep Q-learning for discrete actions \citep{Q_learning} and deep deterministic policy gradient (DDPG) for continuous actions \citep{DDPG}. 
%Algorithm S1 in Appendix 
Algorithm~\ref{Algo: policy2} in Appendix~\ref{appendix:policy-learning} 
gives the detailed procedure of model-free policy learning with ASRs in partially observable environments.

\vspace{-2mm}
\paragraph{Model-Based Policy Learning. }
The downside of model-free RL algorithms is that they are usually data hungry, requiring very large amounts of interactions. On the contrary, model-based RL algorithms enjoy much better sample efficiency. Hence, we make use of the learned generative environment model, including the transition dynamics, observation function, and reward function, for model-based policy optimization. Based on the generative environment model, one can learn behaviors from imagined outcomes to increase sample-efficiency and mitigate heavy and possibly risky interactions with the environment. 
We present the procedure of the classic Dyna algorithm \citep{sutton1990integrated,sutton2018reinforcement} with ASRs in Algorithm~\ref{Algo: policy} in Appendix~\ref{appendix:policy-learning}. %and apply it . 
%When conducting the experiments (Section~\ref{sec:experiment}), we use the classic Dyna algorithm \citep{sutton1990integrated,sutton2018reinforcement} with more details given in the Appendix. 

%because 1) our formulation aims to faithfully achieve the properties that ASRs are expected to exhibit (and only those properties), according to the causal process underlying the RL problem, 

% \textcolor{red}{Need to confirm it with Chaochao}
% Although it is flexible to use different model-based RL algorithms, for a fair comparison with state-of-the-art method--DreamerV2 \citep{DreamerV2}, we use the same actor-critic learning for learning a policy from trajectories imagined in the compact latent space. %Specifically, the action model implements the policy and aims to predict actions that solve the imagination environment. The value model estimates the expected imagined rewards that the action model achieves from each state $\vec{s}^{\,\text{ASR}}_t$. 
% For details, please refer to Algorithm [] in Appendix or \citep{DreamerV2}.

% \vspace{-1mm}
\section{Experiments}
\label{sec:experiment}
% \vspace{-.5mm}

\begin{figure}[htp!]
\centering
\begin{minipage}[t]{.22\textwidth}
  \centering
  \includegraphics[width=0.85\linewidth]{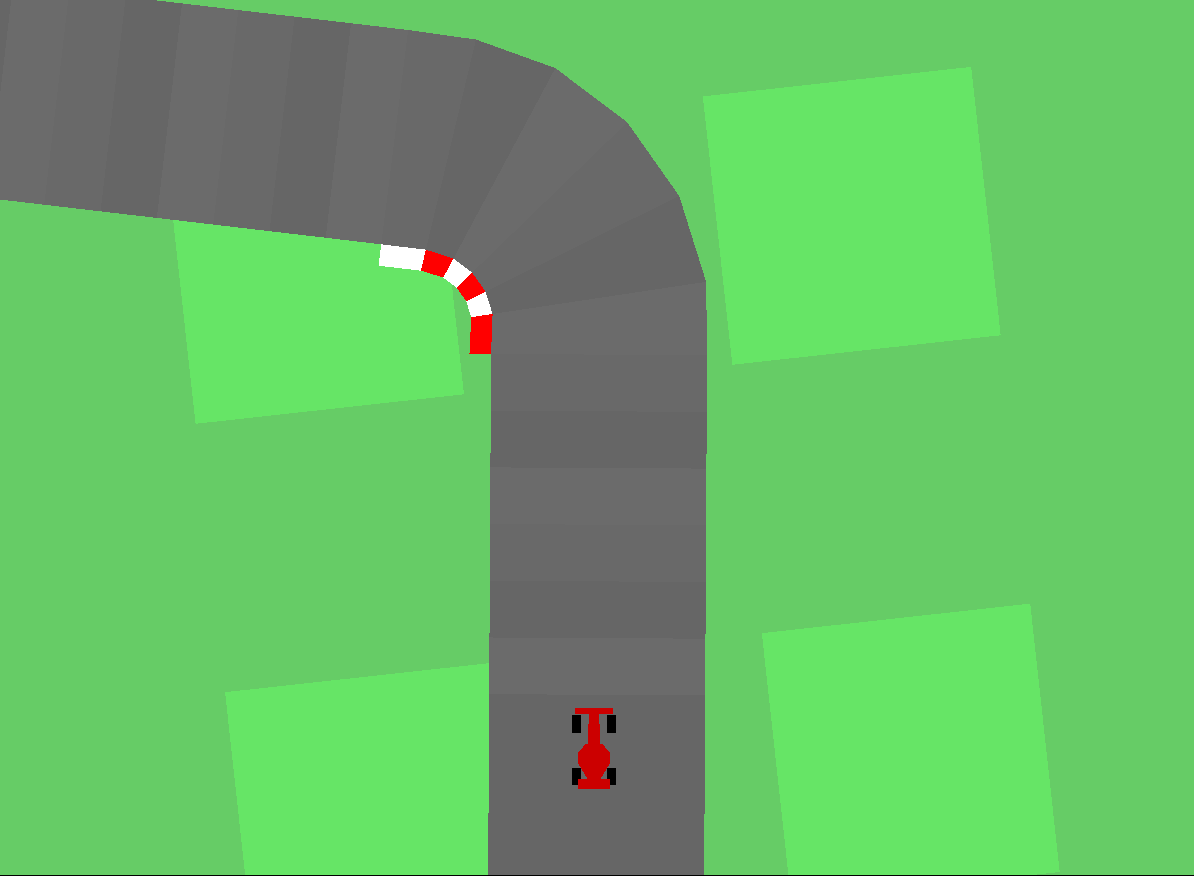}
  \caption{An illustration of Car Racing environment.}
  \label{fig:carracing_image}
\end{minipage}
\hfill
\begin{minipage}[t]{.22\textwidth}
  \centering
  \includegraphics[width=0.9\linewidth]{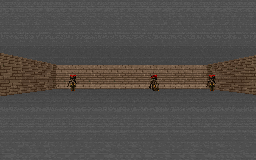}
  \caption{An illustration of VizDoom \textit{take cover} scenario.}
  \label{fig:vizdoom}
\end{minipage}%
\end{figure}

To evaluate the proposed approach, we conducted experiments on both CarRacing environment \citep{Klimov_Carracing} with an illustration in Figure \ref{fig:carracing_image} and VizDoom \citep{Kempka2016ViZDoom} environment with an illustration in Figure \ref{fig:vizdoom}, following the setup in the world model \citep{World_Model} for a fair comparison. It is known that CarRacing is very challenging---the recent world model \citep{World_Model} is the first known solution to achieve the score required to solve the task. Without stated otherwise, all results were averaged across five random seeds, with standard deviation shown in the shaded area.

% \vspace{-1mm}
\subsection{CarRacing Experiment}
% \vspace{-1.5mm}

CarRacing is a continuous control task with three continuous actions: steering left/right, acceleration, and brake. Reward is $-0.1$ every frame and $+1000/N$ for every track tile visited, where $N$ is the total number of tiles in track. It is obvious that the CarRacing environment is partially observable: by just looking at the current frame, although we can tell the position of the car, we know neither its direction nor velocity that are essential for controlling the car. 
For a fair comparison, we followed a similar setting as in \citet{World_Model}. Specifically, we collected a dataset of $10k$ random rollouts of the environment, and each runs with a random policy until failure. The dimensionality of latent states $\tilde{\vec{s}}_t$ was set to $\tilde{d}=32$, determined by hyperparameter tuning.
% \begin{figure}[htp!]
% \centering
%   \includegraphics[width=0.32\linewidth]{figs/carracing_demo.png}
%   \caption{An illustration of Car Racing environment.}
% \label{fig:carracing_image}
% \end{figure}

% For a fair comparison, we followed the same setting as in \citet{World_Model}. Specifically, we collected a dataset of $10k$ random rollouts of the environment, each consisting of $1000$ time steps. The dimensionality of latent states $\tilde{\vec{s}}_t$ was set to $\tilde{d}=32$, and regularization parameters was set to $\lambda_1\!=\!1$, $\lambda_2\!=\!6$, $\lambda_3\!=\!10$, $\lambda_4\!=\!0.1$, and $\lambda\!=\!\beta\!=\!1$, which are determined by hyperparameter turning. 

% More details are placed in the appendices. 
%The mean cumulative rewards together with their standard deviations reported in Figure~\ref{fig:asrs_comparison} 
%are obtained over \text{blue}{how many} runs.
\begin{figure*}[t]
    \centering
             	\setlength{\abovecaptionskip}{0pt}
    \includegraphics[width=0.9\textwidth]{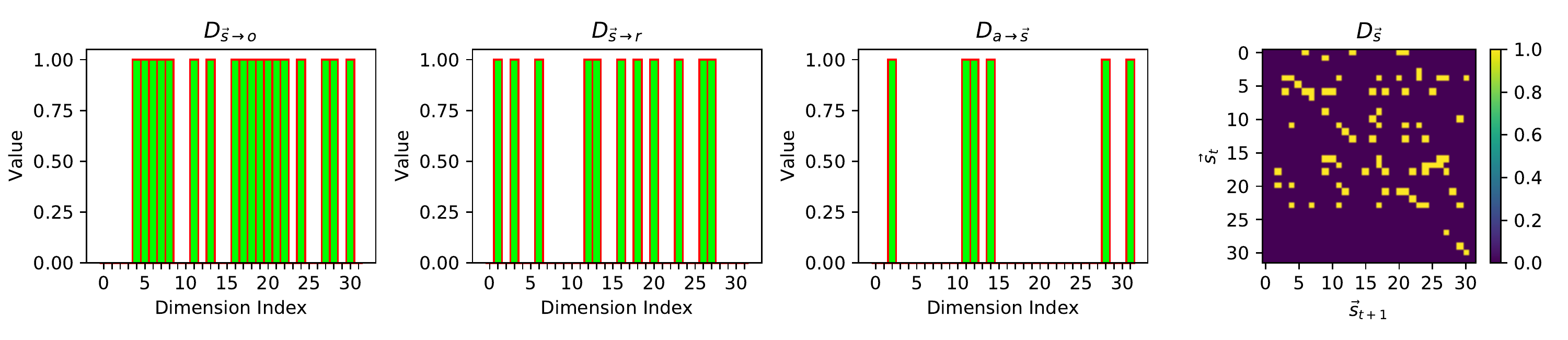} 
    \caption{\small Visualization of estimated structural matrices $D_{\vec{s}\scriptveryshortarrow o}$, $D_{\vec{s} \scriptveryshortarrow r}$, $D_{a \scriptveryshortarrow \vec{s}}$, and $D_{\vec{s}}$ in Car Racing.}
    \label{Fig:asr_analysis}
\end{figure*}

% \vspace{-2mm}
\paragraph{Analysis of ASRs.}
To demonstrate the structures over observed frames, latent states, actions, and rewards, we visualized the learned $D_{\vec{s}\scriptveryshortarrow o}$, $D_{\vec{s} \scriptveryshortarrow r}$, $D_{\vec{s}}$, and $D_{a \scriptveryshortarrow \vec{s}}$, as shown in Figure \ref{Fig:asr_analysis}. Intuitively, we can see that $D_{\vec{s} \scriptveryshortarrow r}$ and $D_{a \scriptveryshortarrow \vec{s}}$ have many values close to zero, meaning that the reward is only influenced by a small number of state dimensions, and not many state dimensions are influenced by the action. Furthermore, from $D_{\vec{s}}$, we found that there are influences from $\tilde{\vec{s}}_{i,t}$ to $\tilde{\vec{s}}_{i,t+1}$ (diagonal values) for most state dimensions, which is reasonable because we want to learn an MDP over the underlying states, while the connections across states (off-diagonal values) are much sparser. Compared to the original 32-dim latent states, ASRs have only 21 dimensions. Below, we empirically showed that the low-dimensional ASRs significantly improve the policy learning performance in terms of both efficiency and efficacy.
% \begin{figure}[t]
%     \centering
%     \includegraphics[width=\textwidth]{figs/asr_analysis_update.pdf} 
%     \caption{Visualization of estimated structural matrices $D_{\vec{s}\scriptveryshortarrow o}$, $D_{\vec{s} \scriptveryshortarrow r}$, $D_{a \scriptveryshortarrow \vec{s}}$, and $D_{\vec{s}}$ in Car Racing.}
%     \label{Fig:asr_analysis}
% \end{figure}

\begin{figure*}[t] 
% \vspace{-2mm}
     \centering
     \begin{minipage}[b]{0.31\linewidth}
         \centering
         \setlength{\abovecaptionskip}{0pt}
         \includegraphics[width=\linewidth]{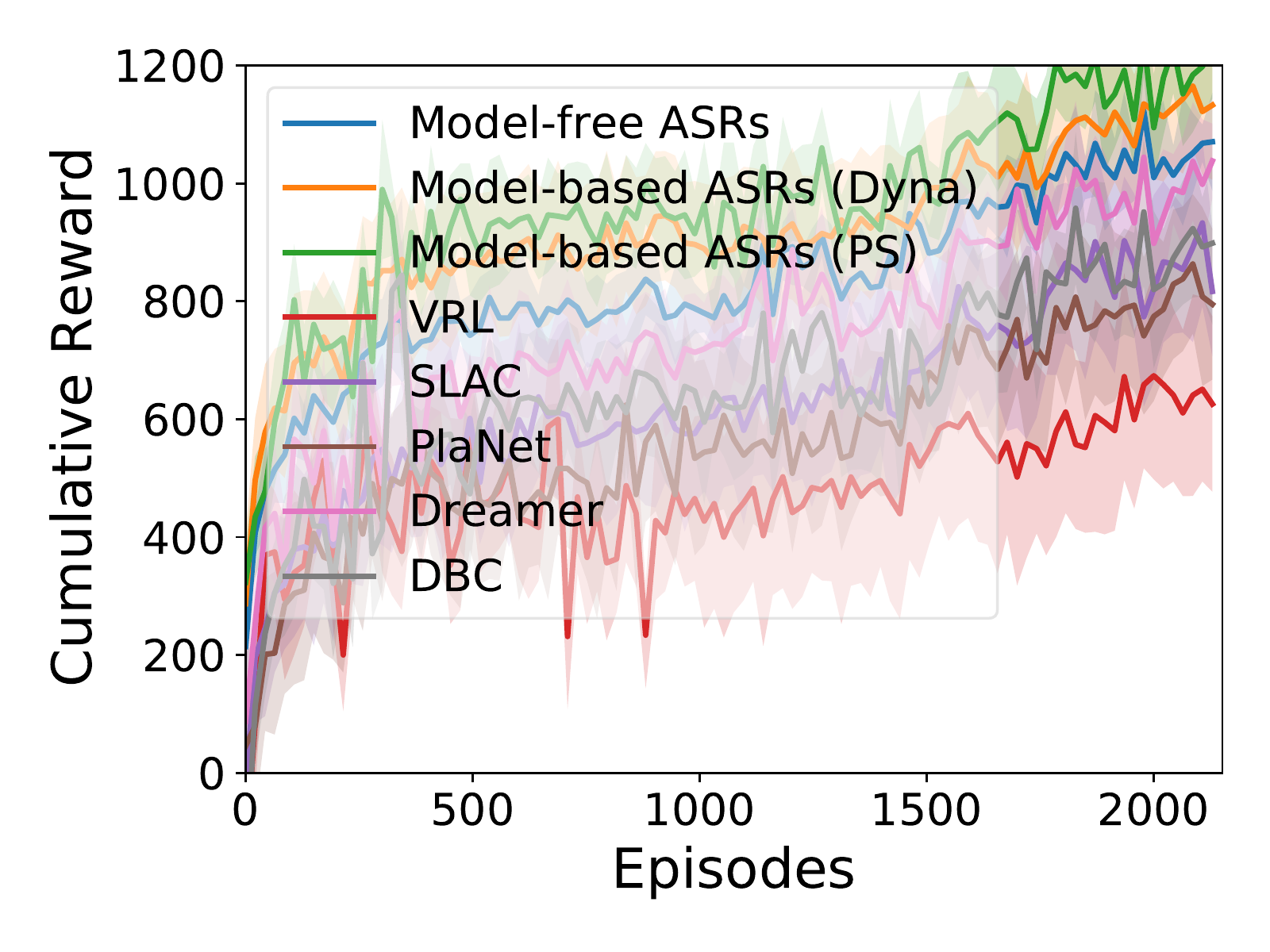}
         \caption{\small Cumulative rewards of model-based ASRs, model-free ASRs, VRL, SLAC, PlaNet, DBC and Dreamer evaluated on CarRacing.} %\vspace{-4mm}
         \label{fig:asrs_comparison}
     \end{minipage} \hspace{4mm}% 
    %  \hfill
     \begin{minipage}[b]{0.31\linewidth}
         \centering
         \setlength{\abovecaptionskip}{0pt}
         \includegraphics[width=\linewidth]{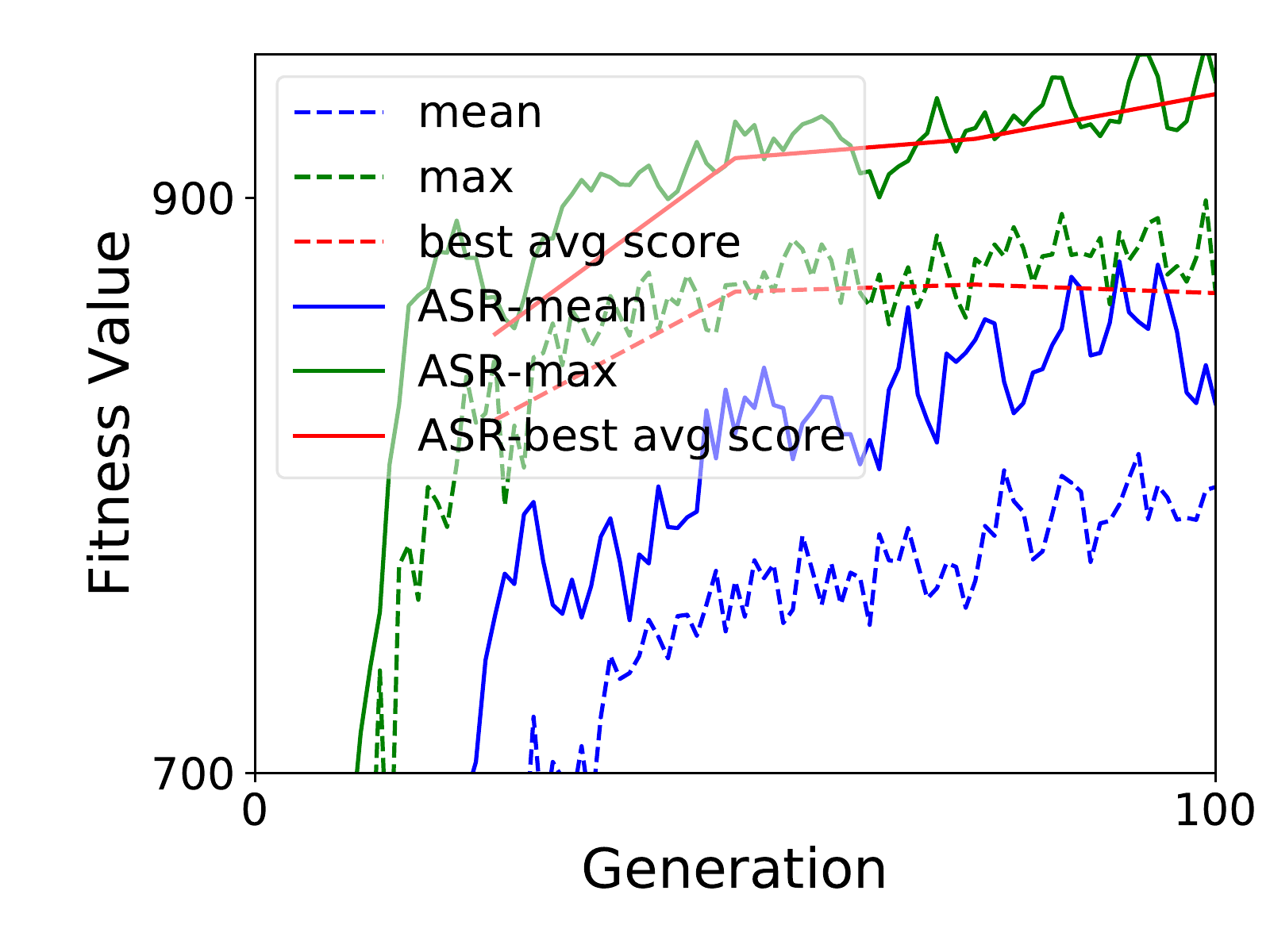}
         \caption{\small Fitness Value of ASRs compared to world models evaluated on CarRacing, including mean score, max score, and the best average score.}%\vspace{-4mm}
         \label{fig:carracing}
     \end{minipage} \hspace{4mm}% 
    %  \hfill
    \begin{minipage}[b]{0.31\linewidth} 
         \centering
         \setlength{\abovecaptionskip}{0pt}
         \includegraphics[width=\linewidth]{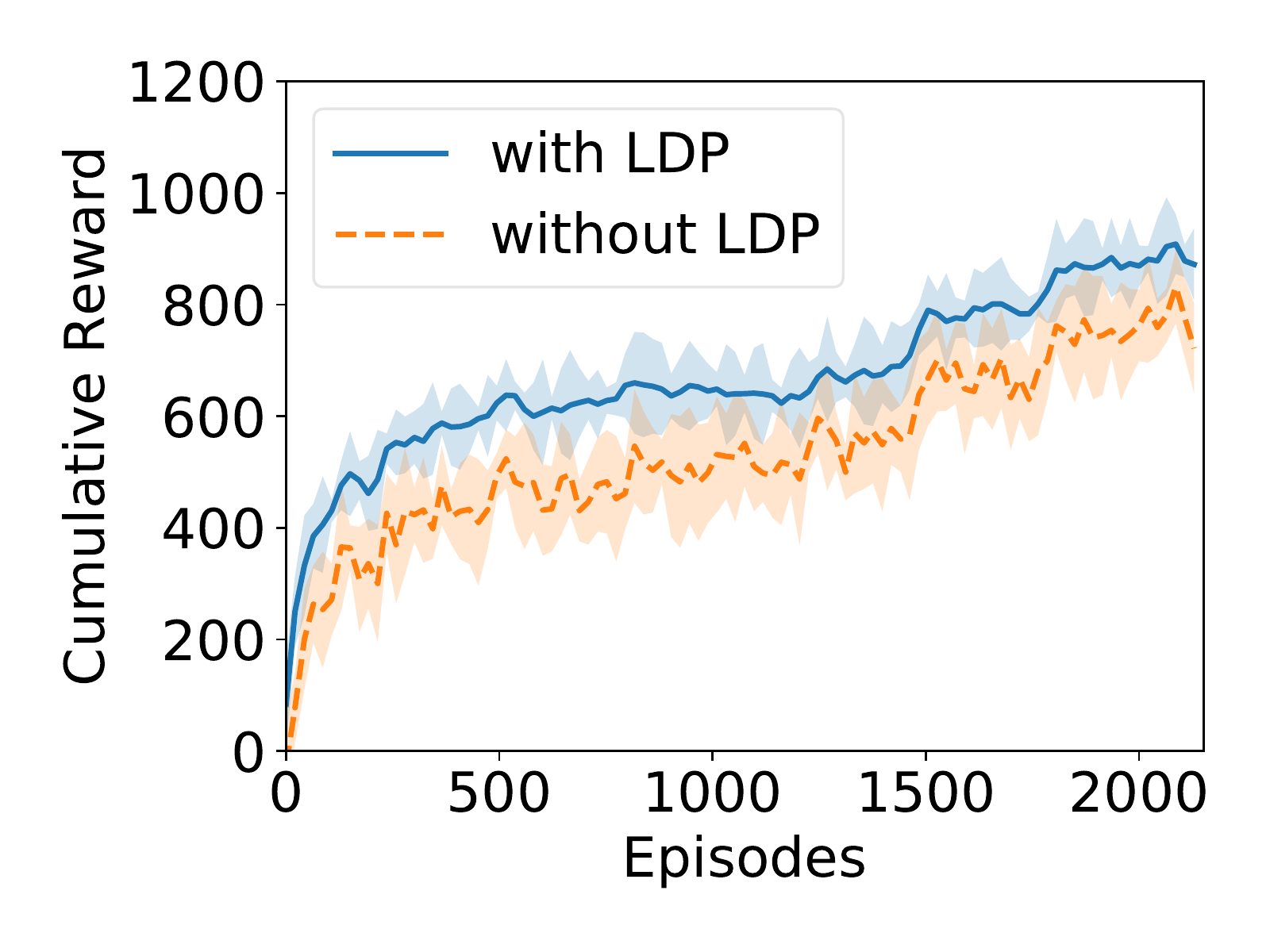}
         \caption{\small Ablation study of latent dynamics prediction (LDP) evaluated on Car Racing with model-free ASR.} \vspace{3.5mm}
         \label{fig:car_ablate}
     \end{minipage}
    %  \vspace{-2mm}
\end{figure*}

% \vspace{-2mm}
\paragraph{Comparison Between Model-Free and Model-Based ASRs.}
% Since the action space is continuous and our goal is to learn Markovian ASRs, we used DDPG \citep{DDPG} for policy learning in this experiment.  %Other RL algorithms that are suitable for both continuous actions and learning Markovian states are also applicable. 
% The ASRs were obtained with different thresholds that were listed in Table \ref{tab:asrs}. 
% As shown in Figure \ref{fig:asrs_comparison}, we can see that 21-dim ASRs (threshold=0.02) have the highest performance even in comparison with 32-dim ASRs (threshold=0). It might be because 32-dim ASRs have redundant or even nuisance information which would mislead the agent. The performance of 15-dim ASRs seems only slightly lower than that of 32-dim ASRs. 
We applied both model-free (DDPG)~\citep{DDPG} and model-based (Dyna and Prioritized Sweeping) algorithms~\citep{sutton1990integrated} to ASRs (with 21-dims). As shown in Figure \ref{fig:asrs_comparison}, interestingly, by taking advantage of the learned generative model, model-based ASRs is superior to model-free ASRs at a faster rate, which demonstrates the effectiveness of the learned model. 
It also shows that with the estimated environment model and ASRs, we can learn behaviors from imagined outcomes to improve sample-efficiency.

% \vspace{-2mm}
\paragraph{Comparison with VRL, SLAC, PlaNet, DBC, and Dreamer.}
We also compared the proposed framework of policy learning with ASRs (with 21-dims) with 1) the same learning strategy but with vanilla representation learning (VRL, implemented without the components for minimal sufficient state representations as in Eq.~\eqref{Suff&Min}), 2) SLAC \citep{SLAC_19}, 3) PlaNet \citep{PlaNet_18}, 4) DBC \citep{Amy_DBM20}, and 5) Dreamer \citep{DreamerV1}. For a fair comparison, the latent dimensions of VRL, PlaNet, SLAC, DBC and Dreamer are set to 21 as well, and we require all of them to have the model capacity similar to ours (i.e., similar model architectures). From Figure \ref{fig:asrs_comparison}, we can see that our methods, both model-free and model-based, obviously outperform others. It is worth noting that the huge performance difference between ASRs and VRL shows that the components for minimal sufficient state representations play a pivotal role in our objective. 

% We also compared ASRs with well-studied sufficient statistics states (VRL), including predictive states \citep{Predictive_state_02, Predictive_state_04}, causal states \citep{Causal_states_19}, and belief states \citep{DVRL}. All these states are sufficient statistic of the history and attempt to be maximally predictive of the future. Here we compare VRL (implemented with predictive states) with the 21-dim ASRs which has the highest performance. For the fair comparison, the dimension of VRL is set to 21 as well. Figure \ref{fig:asrs_comparison} shows the comparison result, where VRL have lower performance. 
% \vspace{-2mm}
\paragraph{Comparison with World Models.}
In light of the fact that world models \citep{World_Model} achieved good performance in CarRacing, we further compared our method (with 21-dim ASRs) with the world model. For a fair comparison, following \citet{World_Model}, we also used the Covariance-Matrix Adaptation Evolution Strategy (CMA-ES) \citep{hansen2016cma} with a population of 64 agents to optimize the parameters of the controller. In addition, following a similar setting as in \citet{World_Model} (where the agent’s fitness value is defined as the average cumulative reward of the 16 random rollouts), we show the fitness values of the best performer (max) and the population (mean) at each generation (Figure \ref{fig:carracing}). 
We also took the best performing agent at the end of every 25 generations and tested it over 1024 random rollout scenarios to record the average (best avg score). It is obvious that our method (denoted by \textit{ASR-*}) has a more efficient and also efficacy training process. The best average score of ASRs is 65 higher than that of world models. 

% \begin{table}[t]
% \centering
% \caption{Comparisons with Dreamer and DBC in CarRacing with natural video distractors, after 2000 training episodes, with standard error.} 
% \label{tab:distractor}
% % \begin{adjustbox}{width=0.98\linewidth,center}
% \begin{tabular}{l|c}
% %\bf X  \\ \bf TRANSFORMER
%  Model & Cumulative Rewards   
% \\ \hline \\
%     Dreamer   &   $621 \pm 124.5$\\
%     DBC        &  $803 \pm 112.5$ \\
%     Model-free ASRs   &  $938 \pm 87.2$ \\
%     Model-based ASRs  &   $954 \pm 98.6$ \\
% \hline                    
% \end{tabular} 
% % \end{adjustbox} 
% \end{table}

\begin{figure*}[t]%\vspace{2mm}
     \centering
     \begin{minipage}{0.3\linewidth}
        \centering
        \setlength{\abovecaptionskip}{12pt}
        \captionsetup{type=table} 
        \begin{adjustbox}{width=\linewidth}
        \begin{tabular}{l D{,}{\pm}{-1} D{,}{\pm}{-1} D{,}{\pm}{4.4}}
        %\bf X  \\ \bf TRANSFORMER 
         Model & \multicolumn{1}{c}{\thead{Cumulative \\ Rewards}} 
        \\ \hline \\
            Dreamer   &   621,124.5\\
            DBC        &  803,112.5 \\
            Model-free ASRs   &  938,87.2 \\
            Model-based ASRs  &   954,98.6 \\
        \hline                    
        \end{tabular} 
        \end{adjustbox}
        \caption{\small Comparisons with Dreamer and DBC in CarRacing with natural video distractors, after 2000 training episodes, with standard error.}  %\vspace{-3mm}
        \label{tab:distractor}
     \end{minipage}\hspace{3mm}
     \begin{minipage}{0.3\linewidth}
         \centering
         \setlength{\abovecaptionskip}{2pt}
         \includegraphics[width=\linewidth]{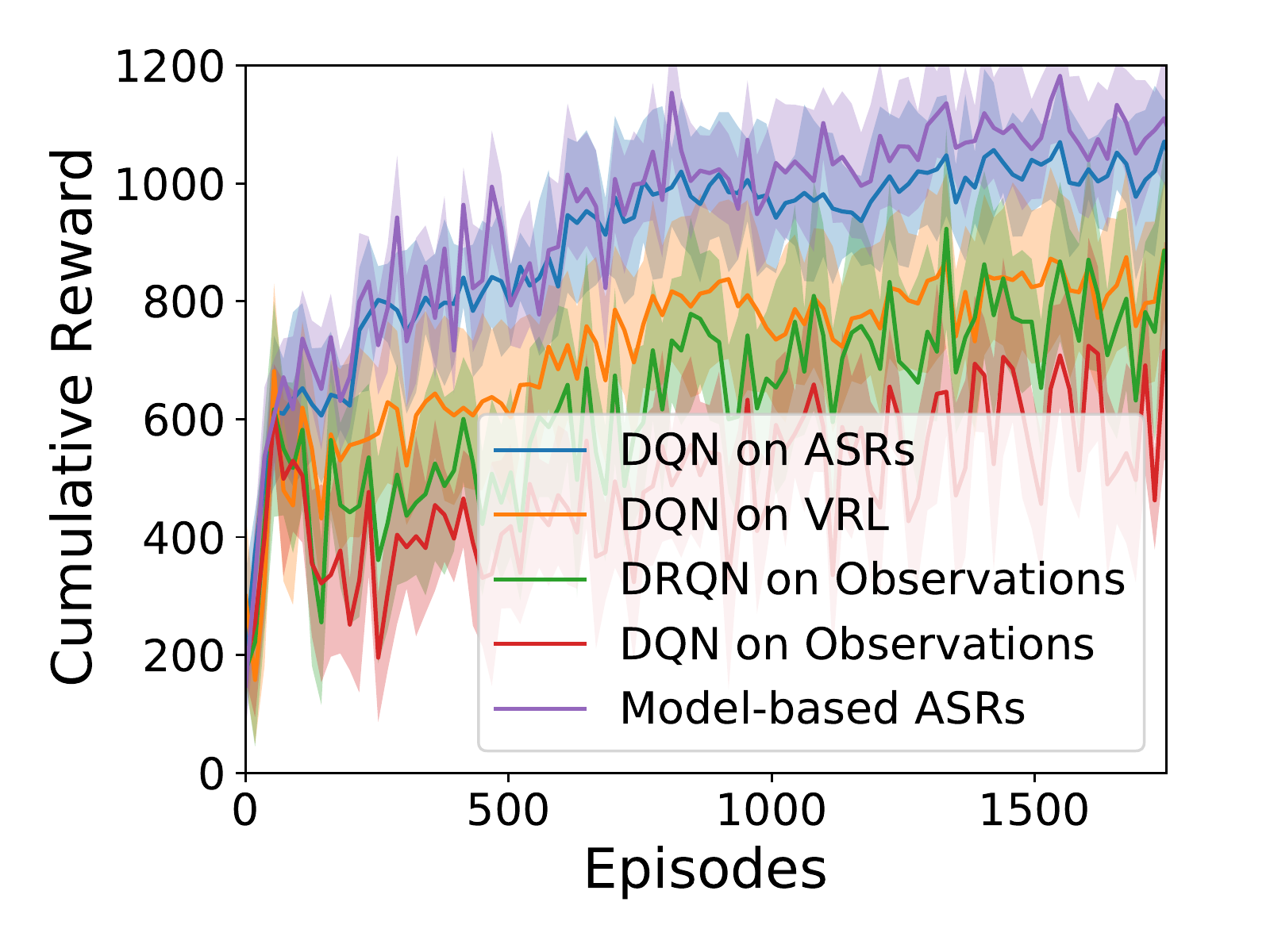} %\vspace{-3mm}
         \caption{\small Comparing ASRs and SOTA methods evaluated on VizDoom.} %\vspace{3mm}
         \label{fig:asrs_vizdoom}
     \end{minipage}\hspace{3mm}
     \begin{minipage}{0.3\linewidth} 
         \centering
         \setlength{\abovecaptionskip}{0pt}
         \includegraphics[width=\linewidth]{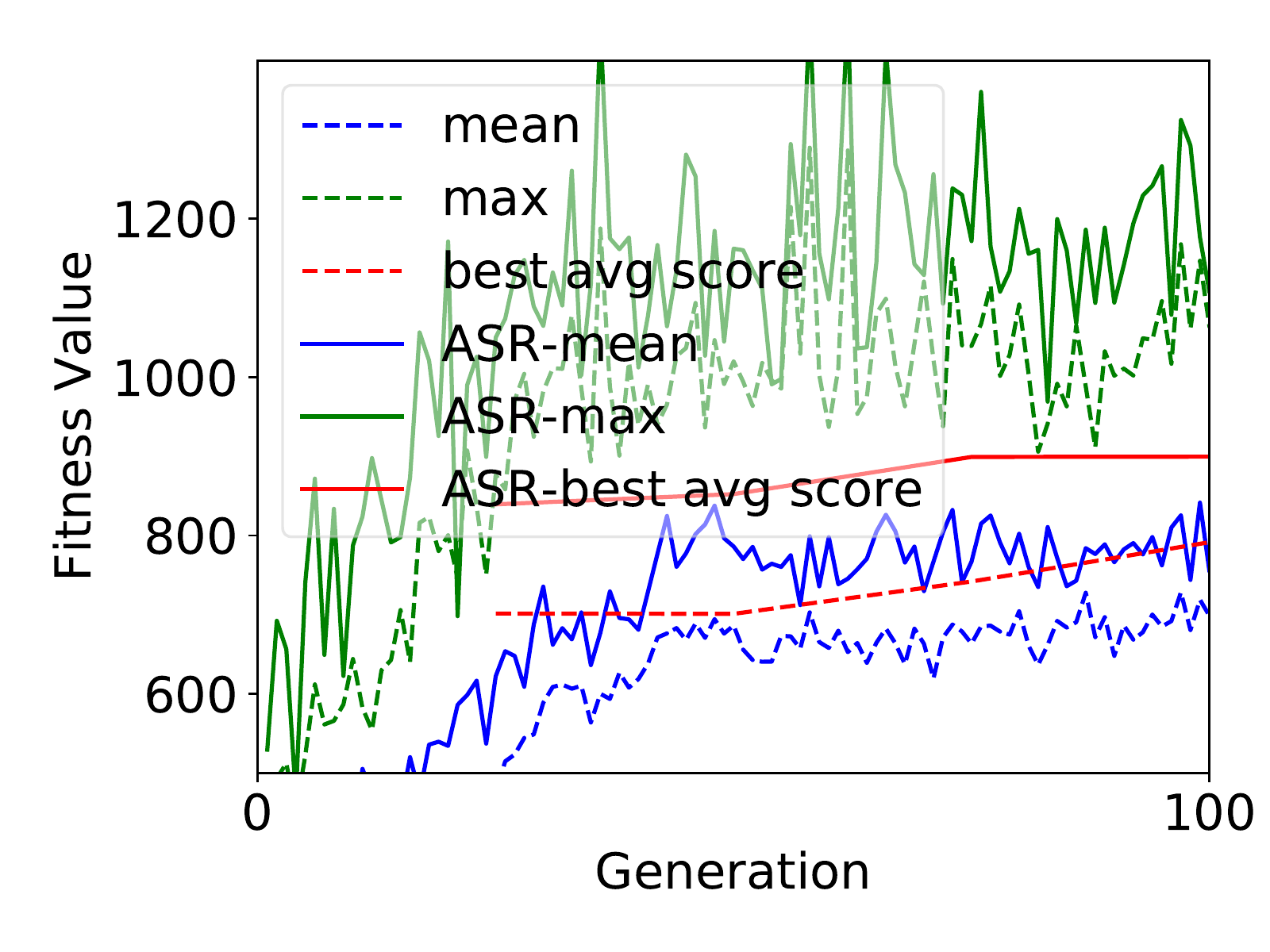}
         \vspace{-3mm}
         \caption{\small Fitness value of ASRs (with CMA-ES) compared to world models evaluated on VizDoom.}
         \label{fig:doom_wm}
     \end{minipage}
     
        % \caption{\small
        % (\ref{fig:asrs_vizdoom}) compares ASRs and SOTA methods evaluated on VizDoom.
        % (\ref{fig:doom_wm}) shows the fitness value of ASRs (with CMA-ES) compared to world models evaluated on VizDoom. 
        % } \vspace{-4mm}
        % \vspace{-2mm}
\end{figure*}

% \vspace{-2mm}
\paragraph{Comparison with Dreamer and DBC with Background Distraction.} We further compared ASRs (with 21-dims) with Dreamer and DBC when there are natural video distractors in CarRacing; we chose Dreamer and DBC, because their performance are relatively better than other comparisons when there are no distractors. Specifically, we followed \citet{Amy_DBM20} to incorporate natural video from the Kinetics dataset \citep{kay2017kinetics} as background in CarRacing. Similarly, for a fair comparison, we require all of them to have the same latent dimensions and have the similar model capacity. As shown in Table \ref{tab:distractor}, we can see that our method outperforms both Dreamer and DBC.

% \vspace{-2mm}
\paragraph{Ablation Study.} We further performed ablation studies on latent dynamics prediction; that is, we compared with the case when the transition dynamics in Fig. \ref{Obj} is not explicitly modeled, but is replaced with a standard normal distribution. Figure \ref{fig:car_ablate} shows that by explicitly modelling the transition dynamics (denoted by \textit{with LDP}), the cumulative reward has an obvious improvement over the one without modelling the transition dynamics (denoted by \textit{without LDP}).

% \begin{figure}[htp!] 
% \centering
% \begin{minipage}{.45\textwidth}
%   \centering
%   \includegraphics[width=\linewidth]{figs/asr_ablation_t0.pdf}
%   \captionof{figure}{Ablation study of latent dynamics prediction (LDP) evaluated on Car Racing.}
%   \label{fig:car_ablate}
% \end{minipage}
% \hfill
% \begin{minipage}{.4\textwidth}
%   \centering
%   \includegraphics[width=\linewidth]{figs/vizdoom_demo.png}
%   \captionof{figure}{An illustration of VizDoom \textit{take cover} scenario.}
%   \label{fig:vizdoom}
% \end{minipage}%
% \end{figure}

% \vspace{-1mm}
\subsection{VizDoom Experiment}
% \vspace{-.5mm}
We also applied the proposed method to VizDoom \textit{take cover} scenario \citep{Kempka2016ViZDoom}, which is a discrete control problem with two actions: move left and move right. Reward is +1 at each time step while alive, and the cumulative reward is defined to be the number of time steps the agent manages to stay alive during an episode. %Therefore, in order to survive as long as possible, the agent has to learn how to avoid fireballs shot by monsters from the other side of the room. In this task, \textit{solving} is defined as attaining the average survival time of greater than 750 time steps over 100 consecutive episodes, each running for a maximum of 2100 time steps.

%We also applied the proposed method to VizDoom \citep{Kempka2016ViZDoom}. VizDoom provides many scenarios and we chose the \textit{take cover} scenario (Figure \ref{fig:vizdoom}). Unlike CarRacing, \textit{take cover} is a discrete control problem with two actions: move left and move right. Reward is +1 at each time step while alive, and the cumulative reward is defined to be the number of time steps the agent manages to stay alive during a episode. Therefore, in order to survive as long as possible, the agent has to learn how to avoid fireballs shot by monsters from the other side of the room. In this task, \textit{solving} is defined as attaining the average survival time of greater than 750 time steps over 100 consecutive episodes, each running for a maximum of 2100 time steps. 

% Following the same setting as in \citep{World_Model}, we collected a dataset of 10k random rollouts of the environment, each consisting of 500 time steps. The dimensionality of latent state $\tilde{\vec{s}}_t$ is set to $\tilde{d}=32$. We also set $\lambda_1=1$, $\lambda_2=6$, $\lambda_3=10$, $\lambda_4=0.1$, and $\lambda=\beta=1$. By tuning thresholds, we finally reported all the results on the  20-dim ASRs, which achieved the best results in all the experiments. 

Considering that in the \textit{take over} scenario the action space is discrete, we applied the widely used DQN \citep{mnih2013playing} on ASRs for policy learning. In addition to the comparisons with VRL (as in CarRacing) and DQN on raw observations, we further compared with another common approach to POMDPs: DRQN \citep{hausknecht2015deep}. As shown in Figure \ref{fig:asrs_vizdoom}, DQN on ASRs achieve a much better performance than all other comparisons, and in particular, DQN on ASRs outperforms DRQN on observations by around 400 on average in terms of cumulative reward.
%In addition to comparing with the results on VRL (as in CarRacing) and a natural comparison with the results on original observations alone, we further compared with another common approach to POMDPs: DRQN \citep{hausknecht2015deep}. As shown in Figure \ref{fig:asrs_vizdoom}, DQN on ASRs achieve a much better performance than all other comparisons, and in particular, DQN on ASRs outperforms DRQN on observations by around 400 on average in terms of cumulative reward.
Similarly, we applied model-based (Dyna) algorithms~\citep{sutton1990integrated} to ASRs (with 21-dims). As shown in Figure \ref{fig:asrs_vizdoom}, we can draw the same conclusion that by taking advantage of the learned generative model, model-based ASRs is superior to model-free ASRs at a faster rate. 
We also applied ASRs to world models, where Figure \ref{fig:doom_wm} shows that our method with ASRs (denoted by \textit{ASR-*}) achieves a better performance. %at a faster rate.

% \vspace{-1.5mm}
\section{Related Work}
% \vspace{-.5mm}
In the past few years, a number of approaches have been proposed to learn low-dimensional Markovian representations, which capture the variation in the environment generated by the agent's actions, without direct supervision \citep{Representation_overview,DKF_15, DVBF_16, World_Model, E2C_15,Solar,Successor_16, Proto_07,Gelada19_ICML,TDVAE-18, Ghosh_ICLR19, Amy_DBM20}. Common strategies for such state representation learning include reconstructing the observation, learning a forward model, or learning an inverse model. Furthermore, prior knowledge, such as temporal continuity \citep{Slowness_02}, can be added to constrain the state space.

Recently, much attention has been paid to world models, which try to learn an abstract representation of both spatial and temporal aspects of the high-dimensional input sequences \citep{E2C_15, Ebert_17, World_Model, PlaNet_18, Planning_Zhang19, Gelada19_ICML, SimPLe, DreamerV1, DreamerV2}. Based on the learned world model, agents can perform model-based RL or planning. Our proposed method is also in the class of world models, which models the generative environment model, and additionally, encodes structural constraints and achieves the sufficiency and minimality of the estimated state representations from the view of generative and selection process. In contrast, \citet{pmlr-v119-shu20a} makes use of contrastive loss, as an alternative of reconstruction loss; however, it only focuses on the transition dynamics and also fails to ensure the sufficiency and minimality.  
Another line of approaches of state representation learning is based on predictive state representations (PSRs) \citep{Predictive_state_02, Predictive_state_04}. %PSRs model a state of controlled dynamical system from a history of performed actions and their resulting observations. %The trait of PSRs is that the predictions are directly related to observable quantities, in contrast to POMDPs, where the state of the system is represented as a probability distribution over unobserved states. 
A recent approach generalizes PSRs to nonlinear predictive models, by exploiting the coarsest partition of histories into classes that are maximally predictive of the future \citep{Causal_states_19}. %In addition, there are works that focus on representation learning in MDP \citep{Successor_16, Proto_07}. 
Moreover, bisimulation-based methods have also attracted much attention \citep{Castro_AAAI20, Amy_DBM20}. 

On the other hand, our work is also related to Bayesian network learning and causal discovery \citep{SGS93, Pearl00, CDNOD}. For example, \citet{factorMDP_07} considers factorized-state MDP with structures being modeled with dynamic Bayesian network or decision trees. Incorporating such structure information has shown benefits in several machine learning tasks \citep{Zhang20_DA, SSM_Huang19}, and in this paper, we show its advantages in POMDPs.
%Incorporating such structure information has also shown benefits in some machine learning tasks, including in domain adaptation \citep{Zhang13_targetshift, Zhang20_DA}, nonstationary time series forecasting \citep{SSM_Huang19}, and clustering \citep{Huang19_NIPS}, and in this paper, we show its advantages in reinforcement learning in partially observable environments.

% \vspace{-1.5mm}
\section{Conclusions and Future Work}
\label{sec:discussion}
% \vspace{-.5mm}
In this paper, we develop a principled framework to characterize a minimal set of state representations that suffice for policy learning, by making use of structural constraints and the goal of maximizing cumulative reward in policy learning. Accordingly, we propose SS-VAE to reliably extract such a set of state representations from raw observations. The estimated environment model and ASRs allow learning behaviors from imagined outcomes in the compact latent space, which effectively reduce sample complexity and possibly risky interactions with the environment. The proposed approach shows promising results on complex environments--CarRacing and Vizdoom. 
%Currently, we only consider the model identifiability in linear cases, while the identifiability in general is nontrivial. 
The future work along this direction include investigating identifiability conditions in general nonlinear cases and extending the approach to cover heterogeneous environments, where the generating processes may change over time or across domains. 
%One of our future work is to investigate identifiability conditions in general nonlinear cases. In addition, in this work, we assume that the environment model is invariant over time. However, in complex real-world scenarios, it may not be the case. Thus, another line of our future work is to extend the approach to cover heterogeneous environments, where the generating processes may change over time or across domains.

\section*{Acknowledgement}
BH would like to acknowledge the support of Apple Scholarship. KZ would like to acknowledge the support by the National Institutes of Health (NIH) under Contract R01HL159805, by the NSF-Convergence Accelerator Track-D award \#2134901, and by a grant from Apple.


\bibliography{reference}
\bibliographystyle{icml2022}

%%%%%%%%%%%%%%%%%%%%%%%%%%%%%%%%%%%%%%%%%%%%%%%%%%%%%%%%%%%%%%%%%%%%%%%%%%%%%%%
%%%%%%%%%%%%%%%%%%%%%%%%%%%%%%%%%%%%%%%%%%%%%%%%%%%%%%%%%%%%%%%%%%%%%%%%%%%%%%%
% APPENDIX
%%%%%%%%%%%%%%%%%%%%%%%%%%%%%%%%%%%%%%%%%%%%%%%%%%%%%%%%%%%%%%%%%%%%%%%%%%%%%%%
%%%%%%%%%%%%%%%%%%%%%%%%%%%%%%%%%%%%%%%%%%%%%%%%%%%%%%%%%%%%%%%%%%%%%%%%%%%%%%%
\newpage
\appendix
\onecolumn

\input{ICML_appendix}

\end{document}

%% file: fig2_new.tex
    \begin{figure}[t]
\centering
\resizebox{.65\linewidth}{.65\linewidth}{
    \begin{tikzpicture}[> = latex, % arrow head style: triangle 45, latex, stealth
    auto,
    observed/.style={circle, draw=black, fill=black!15, thick, inner sep=0pt, minimum size=9mm},
    unobserved/.style={circle, draw=black, thick, inner sep=0pt, minimum size=9mm},
]        

        \node[observed] (o1) { $o_{t-1}$};
        \node[observed] (o2) [right=2cm of o1] { $o_t$};
        \node[observed] (o3) [right=2cm of o2] { $o_{t+1}$};
        \node[unobserved] (s11) [below=0.4cm of o1] { $s_{1,t-1}$};
        \node[unobserved] (s12) [right=2cm of s11] { $s_{1,t}$};
        \node[unobserved] (s13) [right=2cm of s12] { $s_{1,t+1}$};
        \node[unobserved] (s21) [below=0.3cm of s11] { $s_{2,t-1}$};
        \node[unobserved] (s22) [right=2cm of s21] { $s_{2,t}$};
        \node[unobserved] (s23) [right=2cm of s22] { $s_{2,t+1}$};
        \node[unobserved] (s31) [below=0.3cm of s21] { $s_{3,t-1}$};
        \node[unobserved] (s32) [right=2cm of s31] { $s_{3,t}$};
        \node[unobserved] (s33) [right=2cm of s32] { $s_{3,t+1}$};
        
        \node[observed] (a1) [below right=0.3cm and 0.4cm of s31] { $a_{t-1}$};
        \node[observed] (a2) [right=2cm of a1] { $a_t$};
        \node[observed] (r1) [below=0.6cm of s32] { $r_t$};
        \node[observed] (r2) [right=2cm of r1] { $r_{t+1}$};
        
        \node[observed] (R) [below=0.6cm of a2] { $R_t$};

        \path[->, very thick, orange] (s11) edge (o1);
        \path[->, very thick, orange] (s12) edge (o2);
        \path[->, very thick, orange] (s13) edge (o3);
        
        \path[->, very thick, orange] (s21) edge [bend right=40] (o1);
        \path[->, very thick, orange] (s22) edge [bend right=40] (o2);
        \path[->, very thick, orange] (s23) edge [bend right=40] (o3);
        
        \path[->, very thick] (s11) edge (s12);
        \path[->, very thick] (s12) edge (s13);
        \path[->, very thick] (s21) edge (s22);
        \path[->, very thick] (s22) edge (s23);
         \path[->, very thick] (s31) edge (s32);
        \path[->, very thick] (s32) edge (s33);
         \path[->, very thick] (s31) edge (s22);
        \path[->, very thick] (s32) edge (s23);
        
        \path[->, very thick, blue] (s21) edge [bend left=15] (r1);
        % \path[->, very thick, blue] (s31) edge [bend left=15] (r1);
        \path[->, very thick, blue] (s22) edge [bend left=15] (r2);
        % \path[->, very thick, blue] (s32) edge [bend left=15] (r2);
        
        \path[->, very thick, red] (a1) edge [bend left=15] (s12);
        \path[->, very thick, red] (a1) edge [bend left=15] (s22);
        \path[->, very thick, red] (a1) edge (r1);
        \path[->, very thick, red] (a2) edge [bend left=15] (s13);
        \path[->, very thick, red] (a2) edge [bend left=15] (s23);
        \path[->, very thick, red] (a2) edge (r2);
        
        \path[->, very thick, green!60!black] (r1) edge (R);
        \path[->, very thick, green!60!black] (r2) edge (R);
        
        \path[->, thick, dashed] (s11) edge [bend left=30] (a1);
        \path[->, thick, dashed] (s21) edge [bend left=20] (a1);
        \path[->, thick, dashed] (s31) edge (a1);
        
        \path[->, thick, dashed] (s12) edge [bend left=30] (a2);
        \path[->, thick, dashed] (s22) edge [bend left=20] (a2);
        \path[->, thick, dashed] (s32) edge (a2);
    
    \end{tikzpicture}
    }
% ~~~~~~~
\caption{A graphical illustration of the generative environment model. Grey nodes denote observed variables and white nodes represent unobserved variables. Here, $a_{t-1}$ does not have an edge to $s_{3,t}$, and only $s_{2,t-1}$ and $s_{3,t-1}$ have edges to $r_t$, and moreover, we take into account the structural relationships among different dimensions of latent states $\vec{s}_t$. The solid lines represent causal relations, 
while the dashed lines represent predictions of $a_t$ from $\vec{s}_t$, which is not causal, and moreover, dashed lines mean that the relations may not exist and they may differ under different policies.  %\textcolor{red}{maybe mention why we do not draw an edge from $s_{i,t}$ to $a_t$? }
%\leqi{I agree. We should explain this.}
} %Note that here we did not draw an edge from $s_{i,t}$ to $a_t$, because it is an inference procedure, instead of generative.} \vspace{-3mm}
% \vspace{-3mm}
\label{Fig: generalmodel}
    \end{figure}

%% file: ICML_appendix.tex
\section*{\large{\textbf{Appendices for ``Action-Sufficient State Representation Learning for Control with Structural Constraints"}}}
\section{Proof of Proposition 1}
We first give the definitions of the Markov condition and the faithfulness assumption, which will be used in the proof.
\begin{definition}[Global Markov Condition \citep{SGS93, Pearl00}]
 The distribution $p$ over a set of variables $\mathbf{V}$ satisfies the global Markov property on graph $G$ if for any partition $(A, B, C)$ such that if $B$ d-separates $A$ from $C$, then $p(A, C | B) = p(A | B) p(C | B)$.
\end{definition}
\begin{definition}[Faithfulness Assumption \citep{SGS93, Pearl00}]
 There are no independencies between variables that are not entailed by the Markov Condition.
\end{definition}

Below, we give the proof of Proposition 1.
\begin{proof}
The proof contains the following three steps.
\begin{itemize}
    \item In step 1, we show that a state dimension $s_{i,t}$ is in ASRs, that is, it has a directed path to $r_{t+\tau}$ and the path does not go through any action variable (which is equivalent to the recursive definition of ASR), if and only if $s_{i,t} \dependent R_{t+1} | a_{t-1:t}, \vec{s}_{t-1}$.
    \item In step 2, we show that for $s_{i,t}$ with $s_{i,t} \dependent R_{t+1} | a_{t-1:t}, \vec{s}_{t-1}$, if and only if $s_{i,t} \dependent R_{t+1} | a_{t-1:t}, \vec{s}_{t-1}^{\text{ASR}}$.
    \item In step 3, we show that ASRs $\vec{s}_{t}^{\text{ASR}}$ are minimal sufficient for policy learning.
\end{itemize}

\paragraph{Step 1:}

We first show that if a state dimension $s_{i,t}$ is in ASRs, then $s_{i,t} \dependent R_{t+1} | a_{t-1:t}, \vec{s}_{t-1}$.

We prove it by contradiction. Suppose that $s_{i,t}$ is independent of $R_{t+1}$ given $a_{t-1:t}$ and $\vec{s}_{t-1}$. Then according to the faithfulness assumption, we can see from the graph that $s_{i,t}$ does not have a directed path to $r_{t+\tau}$, which contradicts to the assumption, because, otherwise, $a_{t-1:t}$ and $\vec{s}_{t-1}$ cannot break the paths between $s_{i,t}$ and $R_{t+1}$ which leads to the dependence.

We next show that if $s_{i,t} \dependent R_{t+1} | a_{t-1:t}, \vec{s}_{t-1}$, then $s_{i,t} \in \vec{s}_t^{\,\text{ASR}}$.

Similarly, by contradiction suppose that $s_{i,t}$ does not have a directed path to $r_{t+\tau}$. From the graph, it is easy to see that $a_{t-1:t}$ and $\vec{s}_{t-1}$ must break the path between $s_{i,t}$ and $R_{t+1}$. According to the Markov assumption, $s_{i,t}$ is independent of $R_{t+1}$ given $a_{t-1:t}$ and $\vec{s}_{t-1}$, which contradicts to the assumption. Since we have a contradiction, it must be that $s_{i,t}$ has a directed path to $r_{t+\tau}$.

\paragraph{Step 2:}

In step 1, we have shown that $s_{i,t} \dependent R_{t+1} | a_{t-1:t}, \vec{s}_{t-1}$, if and only if it has a directed path to $r_{t+\tau}$.  From the graph, it is easy to see that for those state dimensions which have a directed path to $r_{t+\tau}$, $a_{t-1:t}$ and  $\vec{s}_{t-1}^{\text{ASR}}$ cannot break the path between $s_{i,t}$ and $R_{t+1}$. Moreover, for those state dimensions which do not have a directed path to $r_{t+\tau}$, $a_{t-1:t}$ and  $\vec{s}_{t-1}^{\text{ASR}}$ are enough to break the path between $s_{i,t}$ and $R_{t+1}$.

Therefore, for $s_{i,t}$ with $s_{i,t} \dependent R_{t+1} | a_{t-1:t}, \vec{s}_{t-1}$, if and only if $s_{i,t} \dependent R_{t+1} | a_{t-1:t}, \vec{s}_{t-1}^{\text{ASR}}$.

\paragraph{Step 3:}

In the previous steps, it has been shown that if a state dimension $s_{i,t}$ is in ASRs, then $s_{i,t} \dependent R_{t+1} | a_{t-1:t}, \vec{s}_{t-1}^{\text{ASR}}$, and if a state dimension $s_{i,t}$ is not in ASRs, then $s_{i,t} \independent R_{t+1} | a_{t-1:t}, \vec{s}_{t-1}^{\text{ASR}}$. This implies that $\vec{s}_{t}^{\text{ASR}}$ are minimal sufficient for policy learning to maximize the future reward.

\end{proof}

\section{Learning the ASRs under a Random Policy}

\begin{figure*}
    \centering
    \includegraphics[width=0.6\textwidth]{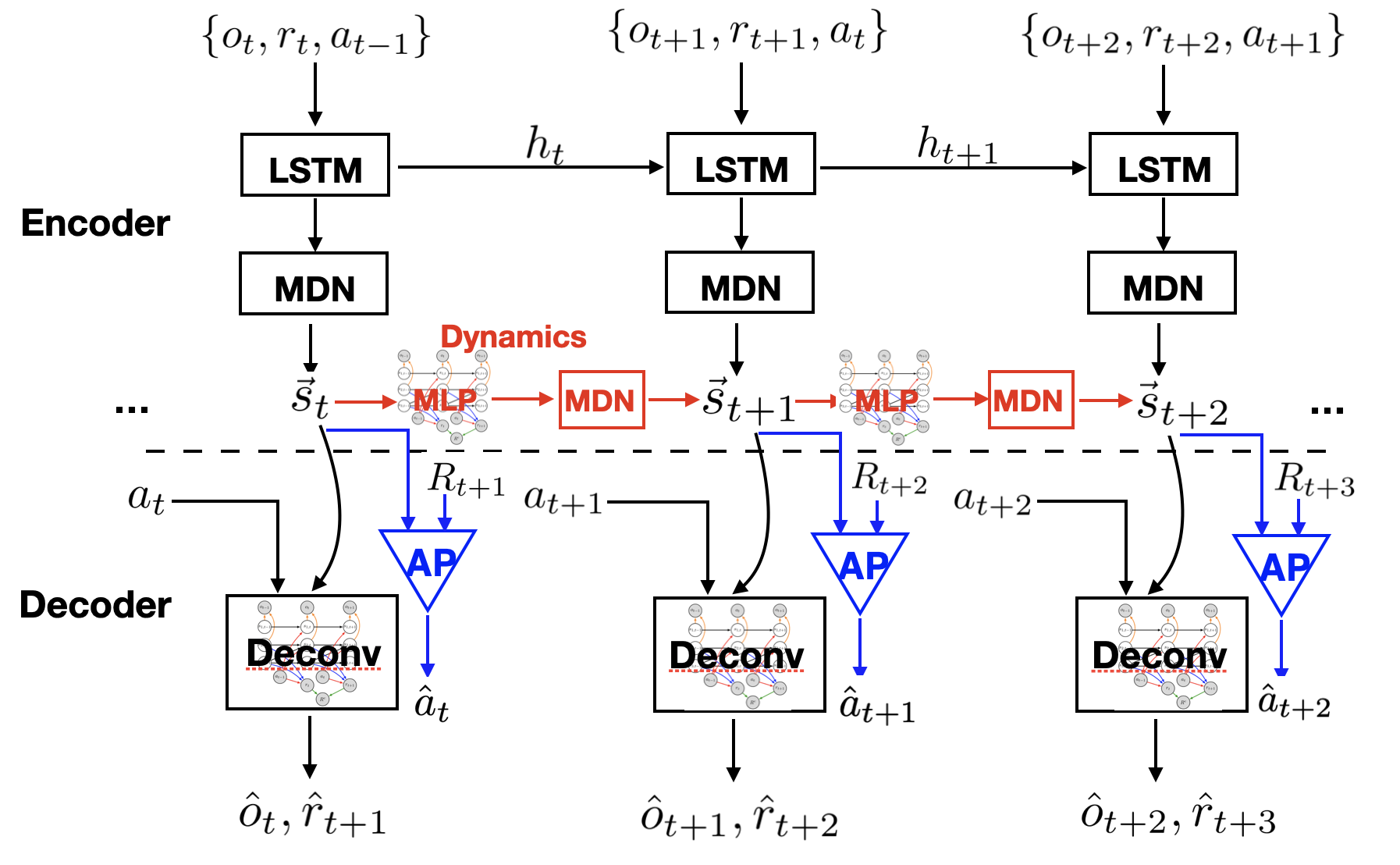}
    \caption{Diagram of neural network architecture to learn state representations. The corresponding structural constraints are involved in ``Deconv'' and ``MLP'', and ``AP" represents the action prediction part for sufficient state representation learning. }
    \label{fig:asr_random_diagram}
\end{figure*}

When collecting the data, that are used to learn the environment model and ASRs, with random actions, it is apparent to observe that actions $a_t$ and ASRs $\vec{s}_t^{\,\text{ASR}}$ are dependent only conditioning on the cumulative reward and previous state---this is a type of dependence relationship induced by selection on the effect (reward). We can then learn the ASRs by maximizing 
\begin{align} \label{eq:asrs_random_policy}
    I(\tilde{\vec{s}}_t^{\,\text{ASR}}; a_t \,|\, R_{t+1},\tilde{\vec{s}}_{t-1}^{\,\text{ASR}})-I(\tilde{\vec{s}}_t^{\,\text{C}}; a_t \,|\, R_{t+1},\tilde{\vec{s}}_{t-1}^{\,\text{ASR}}),
\end{align}
where $\tilde{\vec{s}}^{\,\text{C}} = \tilde{\vec{s}} \backslash \tilde{\vec{s}}^{\,\text{ASR}}$, and $I$ denotes mutual information. Since  $I(\tilde{\vec{s}}_t^{\,\text{ASR}}; a_t \,|\, R_{t+1},\tilde{\vec{s}}_{t-1}^{\,\text{ASR}}) = H(a_t \,|\, R_{t+1},\tilde{\vec{s}}_{t-1}^{\,\text{ASR}}) - H(a_t \,|\, \tilde{\vec{s}}_{t-1:t}^{\,\text{ASR}}, R_{t+1})$, where $H(\cdot)$ denotes the conditional entropy, we can estimate ASRs by maximizing $H(a_t \,|\, R_{t+1},\tilde{\vec{s}}_{t-1}^{\,\text{ASR}}) - H(a_t \,|\, \tilde{\vec{s}}_{t-1:t}^{\,\text{ASR}}, R_{t+1})$, with
\begin{align}
     & H(a_t | \tilde{\vec{s}}_{t-1:t}^{\text{ASR}}, R_{t+1}) = - \mathbb{E}_{q_{\phi,\alpha_1}}  \big\{ \log p_{\alpha_1}( a_t |\tilde{\vec{s}}_{t-1:t}^{\,\text{ASR}}\,, R_{t+1}) \big \}  
     = - \mathbb{E}_{q_{\phi,\alpha_1}} \big\{\log p_{\alpha_1}( a_t |\tilde{D}^{ASR} \odot \tilde{\vec{s}}_{t-1:t}, R_{t+1}) \big \}, \nonumber
\end{align}
and
\begin{align}
     & H(a_t | R_{t+1},\tilde{\vec{s}}_{t-1}^{\,\text{ASR}}) = - \mathbb{E}_{q_{\phi,\alpha_2}}  \big\{ \log p_{\alpha_2}( a_t | R_{t+1},\tilde{\vec{s}}_{t-1}^{\,\text{ASR}}) \big \},  
    = - \mathbb{E}_{q_{\phi,\alpha_2}} \big\{\log p_{\alpha_2}( a_t |\tilde{D}^{ASR} \odot \tilde{\vec{s}}_{t-1}^{\,\text{ASR}}, R_{t+1}) \big \}, \nonumber
\end{align}
where $p_{\alpha_i}$, for $i=1,2$, denotes the probabilistic predictive model of $a_t$ with parameters $\alpha_i$, $q_{\phi\!,\alpha_i}$ is the joint distribution over $\tilde{\vec{s}}_t$ and $a_t$ with $q_{\phi\!,\alpha_i} = q_{\phi}p_{\alpha_i}$, and $q_{\phi}(\tilde{\vec{s}}_t | \tilde{\vec{s}}_{t-1}, \mathbf{y}_{1:t},a_{1:t-1})$ is the probabilistic inference model of $\tilde{\vec{s}}_t$ with parameters $\phi$ and $\mathbf{y}_t \!= \!(o_t^T, r_t^T)$, and $\tilde{D}^{\,\text{ASR}} \in \{0,1\}^{\tilde{d} \times 1}$ is a binary vector indicating which dimensions of $\tilde{\vec{s}}_t$ are in $\tilde{\vec{s}}_t^{\,\text{ASR}}$, so $\tilde{D}^{\,\text{ASR}} \odot \tilde{\vec{s}}_t$ gives ASRs $\tilde{\vec{s}}_t^{\,\text{ASR}}$. Similarly, we can also represent $I(\tilde{\vec{s}}_t^{\,\text{C}}; a_t \,|\, R_{t+1},\tilde{\vec{s}}_{t-1}^{\,\text{ASR}})$ in the same way. Accordingly, the ``sufficiency \& Minimality'' term in the objective function as shown in Figure \ref{Obj} should be replaced by \eqref{eq:asrs_random_policy}, and the corresponding diagram of neural network architecture is as shown in Figure \ref{fig:asr_random_diagram}.

\section{Further Regularization of Minimality}
In this section, we give the detailed derivation on the further regularization of minimality of state representations given in Section 2.1, which is similar to that in the information bottleneck. We achieve it by minimizing conditional mutual information between observed high-dimensional signals $\mathbf{y}_t$, where $\mathbf{y}_t = \{o_t^T, r_t^T \}$, and the ASR $\tilde{\vec{s}}_t^{\,\text{ASR}}$ at time $t$ given data at previous time instances, and meanwhile minimizing the dimensionality of ASRs with sparsity constraints:
 \begin{equation*}
 \setlength{\abovedisplayskip}{3pt}
\setlength{\belowdisplayskip}{3pt}
     \lambda_1 \sum\nolimits_{t=2}^T I(\mathbf{y}_t; \tilde{\vec{s}}_t^{\,\text{ASR}} | \mathbf{y}_{1:t-1}, a_{1:t-1}, \tilde{\vec{s}}_{t-1}) + \lambda_2 \|\tilde{D}^{\,\text{ASR}}\|_1.
 \end{equation*}
 
Note that in the above conditional mutual information, we need to conditional on the previous states $ \tilde{\vec{s}}_{t-1}$, instead of $\tilde{\vec{s}}_{t-1}^{\,\text{ASR}}$, which two give different conditional mutual information. It can be shown by contradiction. Suppose $I(\mathbf{y}_t; \tilde{\vec{s}}_t^{\,\text{ASR}} | \mathbf{y}_{1:t-1}, a_{1:t-1}, \tilde{\vec{s}}_{t-1}) = I(\mathbf{y}_t; \tilde{\vec{s}}_t^{\,\text{ASR}} | \mathbf{y}_{1:t-1}, a_{1:t-1}, \tilde{\vec{s}}_{t-1}^{\,\text{ASR}})$, and denote $\tilde{\vec{s}}^{\,C} = \tilde{\vec{s}} \backslash \tilde{\vec{s}}^{\,\text{ASR}}$. Then the equivalence implies that $\tilde{\vec{s}}_{t-1}^{\,C}$ is independent of $o_t$ (where $o_t \in \mathbf{y}_t$) given $\{\mathbf{y}_{1:t-1}, a_{1:t-1}, \tilde{\vec{s}}_{t-1}^{\,\text{ASR}}\}$. It is obviously violated for the example given in Figure 1, where $\tilde{\vec{s}}^{\,C} = s_1$ and $\tilde{\vec{s}}^{\,\text{ASR}} = \{s_2, s_3\}$, and $s_{1,t-1}$ is dependent on $o_t$ given $\{\mathbf{y}_{1:t-1}, a_{1:t-1}, s_{2,{t-1}}, s_{3,{t-1}}\}$. Hence, conditioning on $ \tilde{\vec{s}}_{t-1}$ and $\tilde{\vec{s}}_{t-1}^{\,\text{ASR}}$ give different conditional mutual information. Therefore, in the above conditional mutual information, we need to condition on the previous states $ \tilde{\vec{s}}_{t-1}$.
 
Moreover, the conditional mutual information $I(\mathbf{y}_t; \tilde{\vec{s}}_t^{\,\text{ASR}} | \mathbf{y}_{1:t-1}, a_{1:t-1}, \tilde{\vec{s}}_{t-1})$ can be upper bound by a KL-divergence, and below we denote $\{\mathbf{y}_{1:t-1}, a_{1:t-1}, \tilde{\vec{s}}_{t-1}\}$ by $\mathbf{z_t}$:
\begin{equation*}
\setlength{\abovedisplayskip}{4pt}
\setlength{\belowdisplayskip}{4pt}
 \begin{array}{lll}
      I(\mathbf{y}_t; \tilde{\vec{s}}_t^{\,\text{ASR}} | \mathbf{y}_{1:t-1}, a_{1:t-1}, \tilde{\vec{s}}_{t-1}) \\
      
      \equiv I(\mathbf{y}_t; \tilde{\vec{s}}_t^{\,\text{ASR}} | \mathbf{z}_t) \\
      
      \equiv \mathbb{E}_{p(\mathbf{y}_t, \tilde{\vec{s}}_t^{\,\text{ASR}},\mathbf{z}_t)} \big\{ \log \frac{q_{\phi}(\tilde{\vec{s}}_t^{\,\text{ASR}}|\mathbf{y}_t, \mathbf{z}_t)}{p(\tilde{\vec{s}}_t^{\,\text{ASR}}| \mathbf{z}_t)} \big\} \\
      
      = \mathbb{E}_{p(\mathbf{y}_t, \tilde{\vec{s}}_t^{\,\text{ASR}},\mathbf{z}_t)} \big\{\log \frac{q_{\phi}(\tilde{\vec{s}}_t^{\,\text{ASR}}|\mathbf{y}_t, \mathbf{z}_t) p_{\gamma}(\tilde{\vec{s}}_t^{\,\text{ASR}}|\tilde{\vec{s}}_{t-1},a_{t-1})}{p(\tilde{\vec{s}}_t^{\,\text{ASR}}| \mathbf{z}_t) p_{\gamma}(\tilde{\vec{s}}_t^{\,\text{ASR}}|\tilde{\vec{s}}_{t-1},a_{t-1})} \big\} \\

      = \mathbb{E}_{p(\mathbf{y}_t, \tilde{\vec{s}}_t^{\,\text{ASR}},\mathbf{z}_t)} \big\{ \log \frac{q_{\phi}(\tilde{\vec{s}}_t^{\,\text{ASR}}|\mathbf{y}_t, \mathbf{z}_t)}{p_{\gamma}(\tilde{\vec{s}}_t^{\,\text{ASR}}|\tilde{\vec{s}}_{t-1},a_{t-1})} \big\} - \mathbb{E}_{p(\mathbf{z}_t)}\big\{\text{KL}(p(\tilde{\vec{s}}_t^{\,\text{ASR}}| \mathbf{z}_t) \Vert p_{\gamma}(\tilde{\vec{s}}_t^{\,\text{ASR}}|\tilde{\vec{s}}_{t-1},a_{t-1}))\big\} \\
      
      \leq \mathbb{E}_{p(\mathbf{y}_t,\mathbf{z}_t)} \big[ \text{KL} \big( q_{\phi}(\tilde{\vec{s}}_t^{\,\text{ASR}}|\mathbf{y}_t, \mathbf{z}_t) \Vert p_{\gamma}(\tilde{\vec{s}}_t^{\,\text{ASR}}|\tilde{\vec{s}}_{t-1},a_{t-1}) \big)\big]\\
      
      \equiv \mathbb{E}_{p(\tilde{\vec{s}}_{t-1}, \mathbf{y}_{1:t},a_{1:t-1})} \big[\text{KL} \big( q_{\phi}(\tilde{\vec{s}}_t^{\,\text{ASR}} | \tilde{\vec{s}}_{t-1}, \mathbf{y}_{1:t},a_{1:t-1}) \Vert p_{\gamma}(\tilde{\vec{s}}_t^{\,\text{ASR}}|\tilde{\vec{s}}_{t-1},a_{t-1}; D_{\vec{s}},D_{a \scriptveryshortarrow \vec{s}}) \big)\big]
 \end{array}
\end{equation*}

with $p_{\gamma}$ being the transition dynamics of $\tilde{\vec{s}}_t$ with parameters $\gamma$.

\section{Assumptions of Proposition 2}
\label{sec:assumption-thm-1}
To show the identifiability of the model in the linear case, we make the following assumptions:
\begin{enumerate}[noitemsep,topsep=0pt]
    \item[A1.] $d_o + d_r \geq d_s$, where $|o_t| = d_o$, $|r_t|=d_r$, and $|s_t|=d_s$.
    \item[A2.] $(D_{\vec{s} \scriptveryshortarrow o}^{\top},D_{\vec{s} \scriptveryshortarrow r}^{\top})$ is full column rank and $D_{\vec{s}}$ is full rank.
    \item[A3.] The control signal $a_t$ is i.i.d. and the state $\vec{s}_t$ is stationary. %That is, the roots of $\det(I-D z^{-1})=0$, where $z$ is the delay operator and $I$ the identity matrix, lie inside the complex unit circle, i.e., with modulus smaller than one.
    \item[A4.] The process noise has a unit variance, i.e., $\text{var}(\eta_t) = I$.
    \end{enumerate}
    
\section{Proof of Proposition 2}
\begin{proof}

The proof of the linear case without control signals has been shown in \citet{zhang2011general}. 
Below, we give the identifiability proof in the linear-Gaussian case with control signals:
\begin{equation}
\left \{
\begin{array}{lll}
o_t = D_{\vec{s} \scriptveryshortarrow o}^{\top} \vec{s}_t +  e_t, \\
r_{t+1} = D_{\vec{s} \scriptveryshortarrow r}^{\top} \vec{s}_{t} + D_{a \scriptveryshortarrow r}^{\top} a_{t} + \epsilon_{t+1}, \\
\vec{s}_t = D_{\vec{s}}^{\top} \vec{s}_{t-1} + D_{a \scriptveryshortarrow \vec{s}}^{\top} a_{t-1} + \eta_t.
\end{array} \right.
\label{Eq: linear}
\end{equation}

Let $\mathbf{y}_{t+1} = [o_t^{\top}, r_{t+1}^{\top}]^{\top}$, $\ddot{D}_{\vec{s} \scriptveryshortarrow o} = [D_{\vec{s} \scriptveryshortarrow o}^{\top}, D_{\vec{s} \scriptveryshortarrow r}^{\top}]^{\top}$, $\ddot{D}_{a \scriptveryshortarrow r} = [\vec{0}^{\top}, D_{a \scriptveryshortarrow r}^{\top}]^{\top}$, and $\ddot{e}_t = [e_t^{\top}, \epsilon_{t+1}^{\top}]^{\top}$. Then the above equation can be represented as:
\begin{equation}
\left \{
\begin{array}{lll}
\mathbf{y}_t = \ddot{D}_{\vec{s} \scriptveryshortarrow o}^{\top} \vec{s}_t + \ddot{D}_{a \scriptveryshortarrow r}^{\top} a_{t} + \ddot{e}_t, \\
\vec{s}_t = D_{\vec{s}}^{\top} \vec{s}_{t-1} + D_{a \scriptveryshortarrow \vec{s}}^{\top} a_{t-1} + \eta_t.
\end{array} \right.
\label{Eq: linear2}
\end{equation}

Because the dynamic system is linear and Gaussian, we make use of the second-order statistics of the observed data to show the identifiability. 
We first consider the cross-covariance between $\mathbf{y}_{t+k}$ and $a_{t}$:
\begin{equation}
\left \{
\begin{array}{lll}
\text{Cov}(\mathbf{y}_{t+k},a_{t}) = \ddot{D}_{\vec{s} \scriptveryshortarrow o}^{\top} D_{\vec{s}}^{k-1} D_{a \scriptveryshortarrow \vec{s}}^{\top} \cdot\text{Var}(a_{t}), & \text{if } k > 0, \\
\text{Cov}(\mathbf{y}_{t+k},a_{t}) = \ddot{D}_{a \scriptveryshortarrow r}^{\top} \cdot \text{Var}(a_{t}), & \text{if } k=0. \\
\end{array} \right.
\label{Eq: mean}
\end{equation}
Thus, from the cross-covariance between $\mathbf{y}_{t+k}$ and $a_{t}$, we can identify $\ddot{D}_{\vec{s} \scriptveryshortarrow o}^{\top} D_{a \scriptveryshortarrow \vec{s}}^{\top}$, $\ddot{D}_{a \scriptveryshortarrow r}$, and $\ddot{D}_{\vec{s} \scriptveryshortarrow o}^{\top} D_{\vec{s}}^{k} D_{a \scriptveryshortarrow \vec{s}}^{\top}$ for $k>0$.

Next, we consider the auto-covariance function of $\vec{s}$. Define the auto-covariance function of $\vec{s}$ at lag $k$ as $\mathbf{R}_{\vec{s}}(k) = \mathbb{E}[\vec{s}_t \vec{s}_{t+k}^{\top}]$, and similarly for $\mathbf{R}_{\mathbf{y}}(k)$. Clearly, $\mathbf{R}_{\vec{s}}(-k) = \mathbf{R}_{\vec{s}}(k)^{\top}$ and $\mathbf{R}_{\mathbf{y}}(-k) = \mathbf{R}_{\mathbf{y}}(k)^{\top}$. Then we have
\begin{equation}
\left \{
\begin{array}{lll}
  \mathbf{R}_{\vec{s}}(k) = \mathbf{R}_{\vec{s}}(k-1)\cdot D_{\vec{s}}, & \text{if } k > 0,\\
  \mathbf{R}_{\vec{s}}(k) = \mathbf{R}_{\vec{s}}^{\top}(1)\cdot D_{\vec{s}} + D_{a \scriptveryshortarrow \vec{s}}^{\top} \text{Var}(a_{t-1}) D_{a \scriptveryshortarrow \vec{s}} + I, & \text{if } k = 0.
\end{array} \right.
\end{equation}
Below, we first consider the case where $d_o+d_r = d_s$. Let $\tilde{\mathbf{y}}_t = \ddot{D}_{\vec{s} \scriptveryshortarrow o}^{\top} \vec{s}_t$, so ${\mathbf{y}}_t = \tilde{\mathbf{y}}_t + \ddot{D}_{a \scriptveryshortarrow r}^{\top} a_{t-1} + \ddot{e}_t$ and $\mathbf{R}_{\tilde{\mathbf{y}}}(k) = \ddot{D}_{\vec{s} \scriptveryshortarrow o}^{\top} \mathbf{R}_{\vec{s}_t}(k) \ddot{D}_{\vec{s} \scriptveryshortarrow o}$. $\mathbf{R}_{\tilde{\mathbf{y}}}(k)$ satisfies the recursive property:
\begin{equation}
\left \{
\begin{array}{lll}
  \mathbf{R}_{\tilde{\mathbf{y}}}(k) = \mathbf{R}_{\tilde{\mathbf{y}}}(k-1) \cdot \Omega^{\top}, & \text{if } k > 0,\\
  \mathbf{R}_{\tilde{\mathbf{y}}}(k) = \mathbf{R}_{\tilde{\mathbf{y}}}^{\top}(1) \cdot \Omega^{\top} + \ddot{D}_{\vec{s} \scriptveryshortarrow o}^{\top} (D_{a \scriptveryshortarrow \vec{s}}^{\top} \text{Var}(a_{t-1})D_{a \scriptveryshortarrow \vec{s}} + I) \ddot{D}_{\vec{s} \scriptveryshortarrow o},  & \text{if } k = 0,
\end{array} \right.
\end{equation}
where $\Omega = \ddot{D}_{\vec{s} \scriptveryshortarrow o}^{\top} D_{\vec{s}} \ddot{D}_{\vec{s} \scriptveryshortarrow o}^{-1}$. 

Denote $S_k = \ddot{D}_{\vec{s} \scriptveryshortarrow o}^{\top} D_{\vec{s}}^{k-1} D_{a \scriptveryshortarrow \vec{s}}^{\top} \cdot \text{Var}(a_{t})$. 
Then we can derive the recursive property for $\mathbf{R}_{\mathbf{y}}(k)$:
\begin{equation*}
\left \{
\begin{array}{lll}
  \mathbf{R}_{\mathbf{y}}(k) = \mathbf{R}_{\mathbf{y}}(k-1) \cdot \Omega^{\top} - \ddot{D}_{a \scriptveryshortarrow r}^{\top} S_{k-1}^{\top} \Omega^{\top} + \ddot{D}_{a \scriptveryshortarrow r}^{\top} S_{k}^{\top}, & \text{if } k >1,\\
  
  \mathbf{R}_{\mathbf{y}}(k) = \mathbf{R}_{\mathbf{y}}(k-1) \cdot \Omega^{\top} - \ddot{D}_{a \scriptveryshortarrow r}^{\top} \text{Var}^{\top}(a_{t}) \ddot{D}_{a \scriptveryshortarrow r} \Omega^{\top} - \Sigma_e \Omega^{\top} + \ddot{D}_{a \scriptveryshortarrow r}^{\top} S_{k}^{\top}, & \text{if } k =1,\\
  \mathbf{R}_{\mathbf{y}}(k) = \mathbf{R}_{\mathbf{y}}^{\top}(1) \cdot \Omega^{\top} + \big (\ddot{D}_{a \scriptveryshortarrow r}^{\top} \text{Var}(a_{t}) \ddot{D}_{a \scriptveryshortarrow r} + \Sigma_e \big) \\~~~~~~~~~~~~~~~+ \ddot{D}_{\vec{s} \scriptveryshortarrow o}^{\top} (D_{a \scriptveryshortarrow \vec{s}}^{\top} \text{Var}(a_{t}) D_{a \scriptveryshortarrow \vec{s}} + I) \ddot{D}_{\vec{s} \scriptveryshortarrow o},  & \text{if } k = 0.
\end{array} \right.
\end{equation*}

When $k=2$, we have $$\mathbf{R}_{\mathbf{y}}(2) = \mathbf{R}_{\mathbf{y}}(1) \cdot \Omega^{\top} -  \ddot{D}_{a \scriptveryshortarrow r}^{\top} S_1^{\top} \Omega^{\top} +  \ddot{D}_{a \scriptveryshortarrow r}^{\top} S_2^{\top}. $$
The above equation can be re-organized as
$$ \big(\mathbf{R}_{\mathbf{y}}(2) - \ddot{D}_{a \scriptveryshortarrow r}^{\top} \cdot S_2^{\top} \big) = \big(\mathbf{R}_{\mathbf{y}}(1) - \ddot{D}_{a \scriptveryshortarrow r}^{\top} \cdot S_1^{\top} \big) \cdot \Omega^{\top}.$$
Because $\ddot{D}_{a \scriptveryshortarrow r}$ and $S_k$ are identifiable, and suppose $\big(\mathbf{R}_{\mathbf{y}}(1) - \ddot{D}_{a \scriptveryshortarrow r}^{\top} \cdot S_1^{\top} \big)$ is invertible, $\Omega = \ddot{D}_{\vec{s} \scriptveryshortarrow o}^{\top} D_{\vec{s}} \ddot{D}_{\vec{s} \scriptveryshortarrow o}^{-1}$ is identifiable.

We further consider $\mathbf{R}_{\mathbf{y}}(0)$ and $\mathbf{R}_{\mathbf{y}}(1)$ and write down the following form:
\begin{equation*}
\begin{array}{lll}
&\begin{bmatrix}
\mathbf{R}_{\mathbf{y}}(0) - \ddot{D}_{\vec{s} \scriptveryshortarrow o}^{\top} (D_{a \scriptveryshortarrow \vec{s}}^{\top} \text{Var}(a_{t-1})D_{a \scriptveryshortarrow \vec{s}} + I) \ddot{D}_{\vec{s} \scriptveryshortarrow o}  \\
\mathbf{R}_{\mathbf{y}}(1) \\
\end{bmatrix}
\\ 
~\\
=&
\begin{bmatrix}
\mathbf{R}^{\top}_{\mathbf{y}}(1)  \\
\mathbf{R}_{\mathbf{y}}(0) \\
\end{bmatrix}
\cdot \Omega^{\top} + 
\begin{bmatrix}
 \ddot{D}_{a \scriptveryshortarrow r}^{\top} \text{Var}(a_{t}) \ddot{D}_{a \scriptveryshortarrow r}  \\
 - \ddot{D}_{a \scriptveryshortarrow r}^{\top} \text{Var}^{\top}(a_{t}) \ddot{D}_{a \scriptveryshortarrow r} \Omega^{\top} + \ddot{D}_{a \scriptveryshortarrow r}^{\top} S_{1}^{\top}\\
\end{bmatrix}  +  \Sigma_e
\begin{bmatrix}
 I \\
 - \Omega^{\top} \\
\end{bmatrix}.
\end{array}
\end{equation*}
From the above two equations we can then identify $\Sigma_e$ and $\ddot{D}_{\vec{s} \scriptveryshortarrow o}^{\top} (D_{a \scriptveryshortarrow \vec{s}}^{\top} \text{Var}(a_{t-1})D_{a \scriptveryshortarrow \vec{s}} + I) \ddot{D}_{\vec{s} \scriptveryshortarrow o}$, and because $\ddot{D}_{\vec{s} \scriptveryshortarrow o}^{\top} D_{a \scriptveryshortarrow \vec{s}}^{\top}$ is identifiable, $\ddot{D}_{\vec{s} \scriptveryshortarrow o}^{\top} \ddot{D}_{\vec{s} \scriptveryshortarrow o}$ is identifiable.

In summary, we have shown the identifiability of $\ddot{D}_{a \scriptveryshortarrow r}$, $\ddot{D}_{\vec{s} \scriptveryshortarrow o}^{\top} D_{a \scriptveryshortarrow \vec{s}}^{\top}$, $\ddot{D}_{\vec{s} \scriptveryshortarrow o}^{\top}  D_{\vec{s}}^k D_{a \scriptveryshortarrow \vec{s}}^{\top}$, $\ddot{D}_{\vec{s} \scriptveryshortarrow o}^{\top} \ddot{D}_{\vec{s} \scriptveryshortarrow o}$, and $\Sigma_e$. 
Furthermore, $\ddot{D}_{\vec{s} \scriptveryshortarrow o}$, $ D_{\vec{s}}$, and $D_{a \scriptveryshortarrow \vec{s}}$ are identified up to some orthogonal transformations. That is, suppose the model in Eq. \ref{Eq: linear} with parameters $({D_{\vec{s} \scriptveryshortarrow o}},{D_{\vec{s} \scriptveryshortarrow r}},D_{a \scriptveryshortarrow r}, { D_{\vec{s}}},{D_{a \scriptveryshortarrow \vec{s}}},\Sigma_e,\Sigma_{\epsilon})$ and that with $(\tilde{D}_{\vec{s} \scriptveryshortarrow o},\tilde{D}_{\vec{s} \scriptveryshortarrow r},\tilde{D}_{a \scriptveryshortarrow r},\tilde{D}_{\vec{s}},\tilde{D}_{a \scriptveryshortarrow \vec{s}},\tilde{\Sigma}_{\tilde{e}},\tilde{\Sigma}_{\tilde{\epsilon}})$ are observationally equivalent, we then have $\tilde{\ddot{D}}_{\vec{s} \scriptveryshortarrow o} = U \ddot{D}_{\vec{s} \scriptveryshortarrow o}$, $\tilde{D}_{a \scriptveryshortarrow r} = D_{a \scriptveryshortarrow r}$, $\tilde{D}_{\vec{s}} = U^{\top} { D_{\vec{s}}} U$, $\tilde{D}_{a \scriptveryshortarrow \vec{s}} ={D_{a \scriptveryshortarrow \vec{s}}}U$, $\tilde{\Sigma}_{\tilde{e}} = \Sigma_e$, and $\tilde{\Sigma}_{\tilde{\epsilon}} = \Sigma_{\epsilon}$, where $U$ is an orthogonal matrix.

Next, we extend the above results to the case where $d_o + d_r > d_s$. Let $\ddot{D}_{\vec{s} \scriptveryshortarrow o (i,\cdot)}$ be the $i$-th row of $\ddot{D}_{\vec{s} \scriptveryshortarrow o}$. Recall that $\ddot{D}_{\vec{s} \scriptveryshortarrow o}^{\top}$ is of full column rank. Then for any $i$, one can show that there always exist $d_s-1$ rows of $\ddot{D}_{\vec{s} \scriptveryshortarrow o}$, such that they, together with $\ddot{D}_{\vec{s} \scriptveryshortarrow o (i,\cdot)}$, form a $d_s \times d_s$ full-rank matrix, denoted by $\bar{\ddot{D}}_{\vec{s} \scriptveryshortarrow o (i,\cdot)}$. Then from the observed data corresponding to $\bar{\ddot{D}}_{\vec{s} \scriptveryshortarrow o (i,\cdot)}$, $\bar{\ddot{D}}_{\vec{s} \scriptveryshortarrow o (i,\cdot)}$ is determined up to orthogonal transformations. Thus, $\ddot{D}_{\vec{s} \scriptveryshortarrow o}$ is identified up to orthogonal transformations. Similarly, $D_{a \scriptveryshortarrow r}$, $D_{\vec{s}}$, and $D_{a \scriptveryshortarrow \vec{s}}$ are identified up to orthogonal transformations. Furthermore, $\text{Cov}(\ddot{D}_{\vec{s} \scriptveryshortarrow o}^{\top} \vec{s}_t + D_{a \scriptveryshortarrow r}^{\top} a_{t})$ is determined by $\ddot{D}_{\vec{s} \scriptveryshortarrow o}$, $\ddot{D}_{a \scriptveryshortarrow r}$, $D_{\vec{s}}$, and $D_{a \scriptveryshortarrow \vec{s}}$. Because $\text{Cov}(\mathbf{y}_t) = \text{Cov}(\ddot{D}_{\vec{s} \scriptveryshortarrow o}^{\top} \vec{s}_t + D_{a \scriptveryshortarrow r}^{\top} a_{t}) + \Sigma_{\ddot{e}}$, $\Sigma_{\ddot{e}}$ is identifiable.

One may further add sparsity constraints on $D_{\vec{s} \scriptveryshortarrow o}$, $D_{\vec{s} \scriptveryshortarrow r}$, $D_{\vec{s}}$, and $D_{a \scriptveryshortarrow \vec{s}}$, to select more sparse structures among the equivalent ones. For example, one may add sparsity constraints on the rows of $D_{\vec{s} \scriptveryshortarrow o}$. Note this corresponds to the mask on the elements of $\vec{s}_t$ in Eq. \ref{Eq: model2}; if the full row is 0, then the corresponding dimension of $\vec{s}_t$ is not selected.

\end{proof}

\section{More Estimation Details for General Nonlinear Models}
\label{sec:estimation-non-linear-model}

The generative model $p_{\theta}$ can be further factorized as follows:
\begin{equation}
    \begin{array}{lll}
      & \log p_{\theta}( \mathbf{y}_{1:T} |\tilde{\vec{s}}_{1:T}, a_{1:T-1}; D_{\vec{s} \scriptveryshortarrow o},D_{\vec{s} \scriptveryshortarrow r},D_{a \scriptveryshortarrow r}) \\
      = & \log p_{\theta}(o_{1:T} |\tilde{\vec{s}}_{1:T}; D_{\vec{s} \scriptveryshortarrow o}) + \log p_{\theta}(r_{1:T} |\tilde{\vec{s}}_{1:T}, a_{1:T-1}; D_{\vec{s} \scriptveryshortarrow r},D_{a \scriptveryshortarrow r}) \\
      = & \sum_{t=1}^T \log p_{\theta}(o_t |\tilde{\vec{s}}_t; D_{\vec{s} \scriptveryshortarrow o}) + \log p_{\theta}(r_t |\tilde{\vec{s}}_{t-1}, a_{t-1}; D_{\vec{s} \scriptveryshortarrow r},D_{a \scriptveryshortarrow r}),
    \end{array}
\end{equation}
where both $p_{\theta}(o_t |\tilde{\vec{s}}_t; D_{\vec{s} \scriptveryshortarrow o})$ and $p_{\theta}(r_t |\tilde{\vec{s}}_{t-1}, a_{t-1}; D_{\vec{s} \scriptveryshortarrow r},D_{a \scriptveryshortarrow r})$ are modelled by mixture of Gaussians, with $D_{\vec{s} \scriptveryshortarrow o}$ indicating the existence of edges from $\tilde{\vec{s}}_t$ to $o_t$ and $D_{\vec{s} \scriptveryshortarrow r}$ indicating the existence of edges from $\tilde{\vec{s}}_{t-1}$ to $r_t$. 

The inference model $q_{\phi}(\tilde{\vec{s}}_{1:T} | \mathbf{y}_{1:T},a_{1:T-1})$ is factorized as
\begin{equation*}
    \begin{array}{lll}
        & \log q_{\phi}(\tilde{\vec{s}}_{1:T} | \mathbf{y}_{1:T},a_{1:T-1}) \\
        = & \log q_{\phi}(\tilde{\vec{s}}_1 | \mathbf{y}_1,a_0) + 
        \sum\limits_{t=2}^T \log q_{\phi}(\tilde{\vec{s}}_t | \tilde{\vec{s}}_{t-1}, \mathbf{y}_{1:t},a_{1:t-1}), \\
       %= & \log q_{\phi}(\vec{s}_1 | \mathbf{y}_{1:T},a_{1:T-1}) + \sum\limits_{t=1}^{\top} \log q_{\phi}(\vec{s}_t | \vec{s}_{t-1}, \mathbf{y}_{t:T},a_{t-1:T-1})
    \end{array}
\end{equation*}
where both $q_{\phi}(\tilde{\vec{s}}_1 | \mathbf{y}_1,a_0)$ and $q_{\phi}(\tilde{\vec{s}}_t | \tilde{\vec{s}}_{t-1}, \mathbf{y}_{1:t},a_{1:t-1})$ are modelled with mixture of Gaussians. %, and they are implemented with LSTM and mixture density network (give more details...).

The transition dynamics $p_{\gamma}$ is factorized as
\begin{equation}
    \begin{array}{lll}
       \log p_{\gamma}(\tilde{\vec{s}}_{1:T} | a_{1:T-1};D_{\vec{s}(\cdot,i)},D_{a \scriptveryshortarrow \vec{s}(\cdot,i)}) 
       = \sum \limits_{t=1}^T \log p_{\gamma}(\tilde{\vec{s}}_t | \tilde{\vec{s}}_{t-1}, a_{t-1};D_{\vec{s}(\cdot,i)},D_{a \scriptveryshortarrow \vec{s}(\cdot,i)}),
    \end{array}
    % \label{Eq: app_dynamics}
\end{equation}
with $\tilde{\vec{s}}_t| \tilde{\vec{s}}_{t-1}$  modelled with mixture of Gaussians.

Thus, the KL divergence can be represented as follows:
\begin{equation}
    \begin{array}{lll}
       & \text{KL} \big( q_{\phi}(\tilde{\vec{s}}_{1:T} | \mathbf{y}_{1:T},a_{1:T-1}) \Vert p_{\gamma}(\tilde{\vec{s}}_{1:T}) \big) \\
       % = & \text{KL} \big( q_{\phi}(\vec{s}_1 | \mathbf{y}_{1:T},a_{1:T-1}) \Vert p_0(\vec{s}_1) \big) \\
       % & + \sum \limits_{t=2}^{\top} \mathbb{E}\limits_{q_{\phi}} \big[ \text{KL} \big( q_{\phi}(\vec{s}_t | \vec{s}_{t-1}, \mathbf{y}_{1:T},a_{1:T-1}) \Vert p_0(\vec{s}_t|\vec{s}_{t-1}) \big) \big]\\
       = & \text{KL} \big( q_{\phi}(\tilde{\vec{s}}_1 | \mathbf{y}_{1},a_0) \Vert p_{\gamma}(\tilde{\vec{s}}_1) \big)  + \sum \limits_{t=2}^T \mathbb{E}_{q_{\phi}} \big[ \text{KL} \big( q_{\phi}(\tilde{\vec{s}}_t | \tilde{\vec{s}}_{t-1}, \mathbf{y}_{1:t},a_{1:t-1}) \Vert p_{\gamma}(\tilde{\vec{s}}_t|\tilde{\vec{s}}_{t-1}) \big) \big].\\
    \end{array}
\end{equation}

In practice, KL divergence with mixture of Gaussians is hard to implement, so instead, we used the following objective function: 
\begin{equation}
    \begin{array}{lll}
       \text{KL} \big( q_{\phi}(\tilde{\vec{s}}_1 | \mathbf{y}_{1},a_0) \Vert p_{\gamma'}(\tilde{\vec{s}}_1) \big)  + \sum \limits_{t=2}^T \mathbb{E}_{q_{\phi}} \big[ \text{KL} \big( q_{\phi}(\tilde{\vec{s}}_t | \tilde{\vec{s}}_{t-1}, \mathbf{y}_{1:t},a_{1:t-1}) \Vert p_{\gamma'}(\tilde{\vec{s}}_t|\tilde{\vec{s}}_{t-1}) \big) \big]\\
       + \lambda \sum \limits_{t=1}^T \log p_{\gamma}(\tilde{\vec{s}}_t | \tilde{\vec{s}}_{t-1}, a_{t-1};D_{\vec{s}(\cdot,i)},D_{a \scriptveryshortarrow \vec{s}(\cdot,i)})
    \end{array}
\end{equation}
where $p_{\gamma'}$ is a standard multivariate Gaussian $\mathcal{N}(\vec{0},I_d)$.

\section{More Details for Policy Learning with ASRs}
\label{appendix:policy-learning}

Algorithm \ref{Algo: policy2} gives the procedure of model-free policy learning with ASRs in partially observable environments. Specifically, it starts from model initialization (line 1) and data collection with a random policy (line 2). Then it updates the environment model and identifies the set of ASRs with the collected data (line 3), after which, the main procedure of policy optimization follows. In particular, because we do not directly observe the states $\vec{s}_t$, on lines 8 and 12, we infer $q_{\phi}(\vec{s}_{t+1}^{\,\text{ASR}} | o_{\leq t+1},r_{\leq t+1},a_{\leq t} )$ and sample $\vec{s}_{t+1}^{\,\text{ASR}}$ from the posterior. The sampled ASRs are then stored in the buffer (line 13). Furthermore, we randomly sample a minibatch of $N$ transitions to optimize the policy (lines 14 and 15). One may perform various RL algorithms on the ASRs, such as deep deterministic policy gradient (DDPG \citep{DDPG}) or Q-learning \citep{Q_learning}.

Algorithm~\ref{Algo: policy} presents the procedure of the classic model-based Dyna algorithm with ASRs. Lines 17-22 make use of the learned environment model to predict the next step, including  $\vec{s}_{t+1}^{\,\text{ASR}}$ and $r_{t+1}$, and update the Q function $n$ times. Specifically, in our implementation, the hyper-parameter $n$ is 20. Based on the learned model, the agent learns behaviors from imagined outcomes in the compact latent space, which helps to increase sample efficiency.

\begin{algorithm}[htp!] 
  \caption{Model-Free Policy Learning with ASRs in Partially Observable Environments}
  \begin{algorithmic}[1]
	\STATE Randomly initialize neural networks and initialize replay buffer $\mathcal{B}$.
	  \STATE Apply random control signals and record multiple rollouts.
	  \STATE Estimate the model given in~\eqref{Eq: model2} with the recorded data (according to Section \ref{Sec: estimation}).
	  \STATE Identify indices of ASRs according to the learned graph structure and the criteria in Prop. \ref{Proposition: ASR}. 
	  \FOR {episode = 1, \ldots, M}
	    \STATE Initialize a random process $\mathcal{N}$ for action exploration.
	    \STATE Receive initial observations $o_1$ and $r_1$.
	    \STATE Infer the posterior $q_{\phi}(\vec{s}_1^{\,\text{ASR}} | o_1,r_1)$ and sample $\vec{s}_1^{\,\text{ASR}}$.
	    \FOR {t = 1, \ldots, T}
	      \STATE Select action $a_t = \pi(\vec{s}_t^{\,\text{ASR}}) + \mathcal{N}_t$ according to the current policy and exploration noise.
	      \STATE Execute action $a_t$ and receive reward $r_{t+1}$ and observation $o_{t+1}$.
	      \STATE Infer the posterior $q_{\phi}(\vec{s}_{t+1}^{\,\text{ASR}} | o_{\leq t+1},r_{\leq t+1},a_{\leq t})$ and sample $\vec{s}_{t+1}^{\,\text{ASR}}$.
	      \STATE Store transition $(\vec{s}_t^{\,\text{ASR}}, a_t, r_{t+1}, \vec{s}_{t+1}^{\,\text{ASR}})$ in $\mathcal{B}$.
	      \STATE Sample a random minibatch of $N$ transitions $(\vec{s}_i^{\,\text{ASR}}, a_i, r_{i+1}, \vec{s}_{i+1}^{\,\text{ASR}})$ from $\mathcal{B}$.
          \STATE Update network parameters using a specified RL algorithm (e.g., DQN or DDPG).
	    \ENDFOR
	  \ENDFOR
  \end{algorithmic}	 
	\label{Algo: policy2}
\end{algorithm}

\begin{algorithm}[htp!] 
  \caption{Model-Based Policy Learning with ASRs in Partially Observable Environments}
  \begin{algorithmic}[1]
	\STATE Randomly initialize neural networks and initialize replay buffer $\mathcal{B}$.
	  \STATE Apply random control signals and record multiple rollouts.
	  \STATE Estimate the model given in~\eqref{Eq: model2} with the recorded data (according to Section \ref{Sec: estimation}).
	  \STATE Identify indices of ASRs according to the learned graph structure and the criteria in Prop. \ref{Proposition: ASR}. 
	  \FOR {episode = 1, \ldots, M}
	    \STATE Initialize a random process $\mathcal{N}$ for action exploration.
	    \STATE Receive initial observations $o_1$ and $r_1$.
	    \STATE Infer the posterior $q_{\phi}(\vec{s}_1^{\,\text{ASR}} | o_1,r_1)$ and sample $\vec{s}_1^{\,\text{ASR}}$.
	    \FOR {t = 1, \ldots, T}
	      \STATE Select action $a_t = \pi(\vec{s}_t^{\,\text{ASR}}) + \mathcal{N}_t$ according to the current policy and exploration noise.
	      \STATE Execute action $a_t$ and receive reward $r_{t+1}$ and observation $o_{t+1}$.
	      \STATE Infer the posterior $q_{\phi}(\vec{s}_{t+1}^{\,\text{ASR}} | o_{\leq t+1},r_{\leq t+1},a_{\leq t})$ and sample $\vec{s}_{t+1}^{\,\text{ASR}}$.
	      \STATE Store transition $(\vec{s}_t^{\,\text{ASR}}, \vec{s}_t, a_t, r_{t+1}, \vec{s}_{t+1}^{\,\text{ASR}}, \vec{s}_{t+1}, o_{t+1})$ in $\mathcal{B}$.
	      \STATE Sample a random minibatch of $N$ transitions $(\vec{s}_i^{\,\text{ASR}}, a_i, r_{i+1}, \vec{s}_{i+1}^{\,\text{ASR}})$ from $\mathcal{B}$.
          \STATE Update network parameters using a specified RL algorithm (e.g., DQN or DDPG).
          \STATE Update the model given in~\eqref{Eq: model2} with the recorded data from $\mathcal{B}$ (according to Section \ref{Sec: estimation}).
          \FOR {p = 1, \ldots, n}
	        \STATE Sample a random minibatch of pairs of $(\vec{s}_t, a_t)$ from $\mathcal{B}$.
	        \STATE Predict $(\vec{s}_{t+1}^{\,\text{ASR}}, r_{t+1})$ according to the model given in~\eqref{Eq: model2}.
	        \STATE Update network parameters using a specified RL algorithm (e.g., DQN or DDPG).
	      \ENDFOR
	    \ENDFOR
	  \ENDFOR
  \end{algorithmic}	 
	\label{Algo: policy}
\end{algorithm}

\section{Additional Experiments and Details}
\label{appendix:additionalExp}

\subsection{CarRacing Experiment}

CarRacing is a continuous control task with three continuous actions: steering left/right, acceleration, and brake. Reward is $-0.1$ every frame and $+1000/N$ for every track tile visited, where $N$ is the total number of tiles in track. It is obvious that the CarRacing environment is partially observable: by just looking at the current frame, although we can tell the position of the car, we know neither its direction nor velocity that are essential for controlling the car.

% \begin{figure}[htp!]
% \centering
%   \includegraphics[width=0.32\linewidth]{figs/carracing_demo.png}
%   \caption{An illustration of Car Racing environment.}
% \label{fig:carracing_image}
% \end{figure}

For a fair comparison, we followed a similar setting as in \citet{World_Model}. Specifically, we collected a dataset of $10k$ random rollouts of the environment, and each runs with random policy until failure, for model estimation. The dimensionality of latent states $\tilde{\vec{s}}_t$ was set to $\tilde{d}=32$, and regularization parameters was set to $\lambda_1\!=\!1$, $\lambda_2\!=\!1$, $\lambda_3\!=\!1$, $\lambda_4\!=\!1$, $\lambda_5\!=\!1$, $\lambda_6\!=\!6$, $\lambda_7\!=\!10$, $\lambda_8\!=\!0.1$, which are determined by hyperparameter tuning. 

\begin{figure}[htp!]
    \centering
    \includegraphics[width=\textwidth]{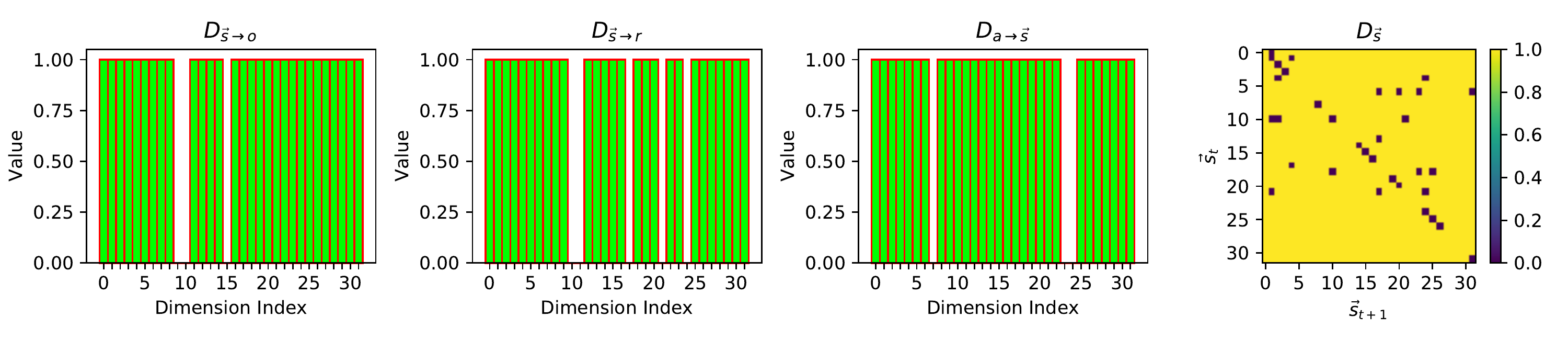} 
    \caption{Visualization of estimated structural matrices $D_{\vec{s}\scriptveryshortarrow o}$, $D_{\vec{s} \scriptveryshortarrow r}$, $D_{a \scriptveryshortarrow \vec{s}}$, and $D_{\vec{s}}$ in Car Racing, without the explicit sparsity constraints.}
    \label{Fig:asr_analysis_wo_constraint}
\end{figure}

\paragraph{Without sparsity constraints.} Figure \ref{Fig:asr_analysis_wo_constraint} gives the estimated structural matrices $D_{\vec{s}\scriptveryshortarrow o}$, $D_{\vec{s} \scriptveryshortarrow r}$, $D_{a \scriptveryshortarrow \vec{s}}$, and $D_{\vec{s}}$ in CarRacing, without the explicit sparsity constraints, where the connections are very dense.

\paragraph{Difference between our SS-VAE and Planet, Dreamer.} Both our method and Planet \citep{PlaNet_18} and Dreamer \citep{DreamerV1} are world model-based methods. The differences are mainly in two aspects: (1) our method explicitly considers the structural relationships among variables in the RL system, and (2) it guarantees minimal sufficient state representations for policy learning. Previous approaches usually fail to take into account whether the extracted state representations are sufficient and necessary for downstream policy learning. Moreover, as for the component of recurrent networks, SS-VAE uses LSTM that only contains the stochastic part, while PlaNet and Dreamer use RSSM that contains both deterministic and stochastic components.

% \begin{figure}[htp!] 
% \centering
% \begin{minipage}{.45\textwidth}
%   \centering
%   \includegraphics[width=\linewidth]{figs/asr_ablation_t0.pdf}
%   \captionof{figure}{Ablation study of latent dynamics prediction (LDP) evaluated on Car Racing with model-free ASR.}
%   \label{fig:car_ablate}
% \end{minipage}
% \hfill
% \begin{minipage}{.4\textwidth}
%   \centering
%   \includegraphics[width=\linewidth]{figs/vizdoom_demo.png}
%   \captionof{figure}{An illustration of VizDoom \textit{take cover} scenario.}
%   \label{fig:vizdoom}
% \end{minipage}%
% \end{figure}

\subsection{VizDoom Experiment}

We also applied the proposed method to VizDoom \citep{Kempka2016ViZDoom}. VizDoom provides many scenarios and we chose the \textit{take cover} scenario. Unlike CarRacing, \textit{take cover} is a discrete control problem with two actions: move left and move right. Reward is +1 at each time step while alive, and the cumulative reward is defined to be the number of time steps the agent manages to stay alive during a episode. Therefore, in order to survive as long as possible, the agent has to learn how to avoid fireballs shot by monsters from the other side of the room. In this task, \textit{solving} is defined as attaining the average survival time of greater than 750 time steps over 100 consecutive episodes, each running for a maximum of 2100 time steps.

Following a similar setting as in \citet{World_Model}, we collected a dataset of 10k random rollouts of the environment, and each runs with random policy until failure. The dimensionality of latent state $\tilde{\vec{s}}_t$ is set to $\tilde{d}=32$. We also set $\lambda_1\!=\!1$, $\lambda_2\!=\!1$, $\lambda_3\!=\!1$, $\lambda_4\!=\!1$, $\lambda_5\!=\!1$, $\lambda_6\!=\!6$, $\lambda_7\!=\!10$, $\lambda_8\!=\!0.1$. By tuning thresholds, we finally reported all the results on the  21-dim ASRs, which achieved the best results in all the experiments. 

\begin{figure}[htp!]
    \centering
    \includegraphics[width=\textwidth]{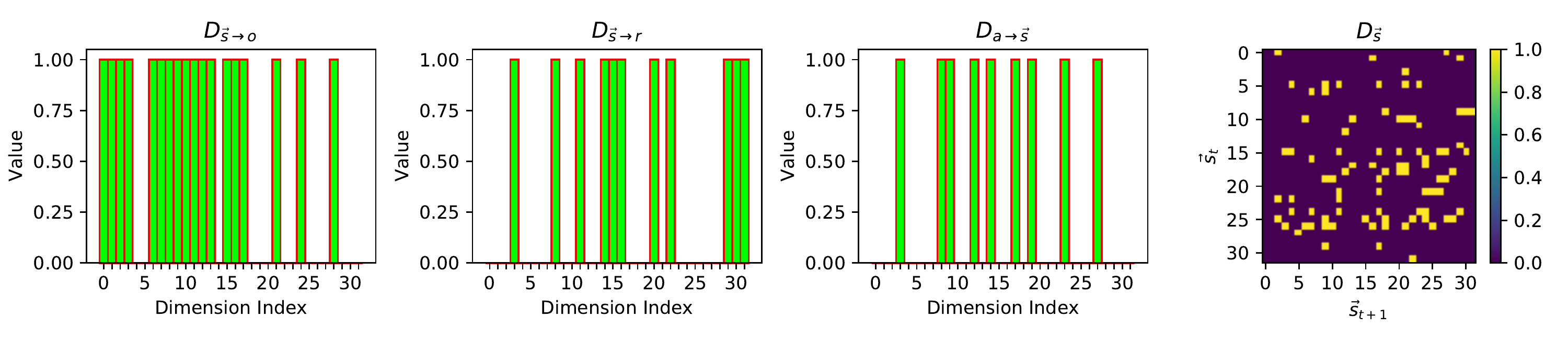} 
    \caption{Visualization of estimated structural matrices $D_{\vec{s}\scriptveryshortarrow o}$, $D_{\vec{s} \scriptveryshortarrow r}$, $D_{a \scriptveryshortarrow \vec{s}}$, and $D_{\vec{s}}$ in VizDoom.}
    \label{Fig:asr_analysis_vizdoom}
\end{figure}

\paragraph{Analysis of ASRs.} 
Similar to the analysis in CarRacing, we also visualized the learned $D_{\vec{s}\scriptveryshortarrow o}$, $D_{\vec{s} \scriptveryshortarrow r}$, $D_{\vec{s}}$, and $D_{a \scriptveryshortarrow \vec{s}}$ in VizDoom, as shown in Figure \ref{Fig:asr_analysis_vizdoom}. Intuitively, we can see that $D_{\vec{s} \scriptveryshortarrow r}$ and $D_{a \scriptveryshortarrow \vec{s}}$ have many values close to zero, meaning that the reward is only influenced by a small number of state dimensions, and not many state dimensions are influenced by the action. Furthermore, from $D_{\vec{s}}$, we found that the connections across states are sparse.

%Considering that in the \textit{take over} scenario the action space is discrete and ASRs are Markovian, we applied the widely used DQN \citep{mnih2013playing} on ASRs for policy learning. In addition to comparing with the results on VRL (as in CarRacing) and a natural comparison with the results on original observations alone, we further compared with another common approach to POMDPs: DRQN \citep{hausknecht2015deep}. As shown in Figure \ref{fig:asrs_vizdoom}, DQN on ASRs achieve a much better performance than all other comparisons, and in particular, DQN on ASRs outperforms DRQN on observations by around 400 on average in terms of cumulative reward.

% Similarly, we applied model-based (Dyna) algorithms~\citep{sutton1990integrated} to ASRs (with 21-dims). As shown in Figure \ref{fig:asrs_vizdoom}, we can draw the same conclusion that by taking advantage of the learned generative model, model-based ASRs is superior to model-free ASRs at a faster rate. 
% We also applied ASRs to world models, where CMA-ES is used to optimise the parameters of the controller. As shown in Figure \ref{fig:doom_wm}, it is evident to see that our method with ASRs (denoted by \textit{ASR-*}) achieves a better performance at a faster rate.

\section{Detailed Model Architectures}
\begin{figure}[htp!]
    \centering
    \includegraphics[width=0.8\linewidth]{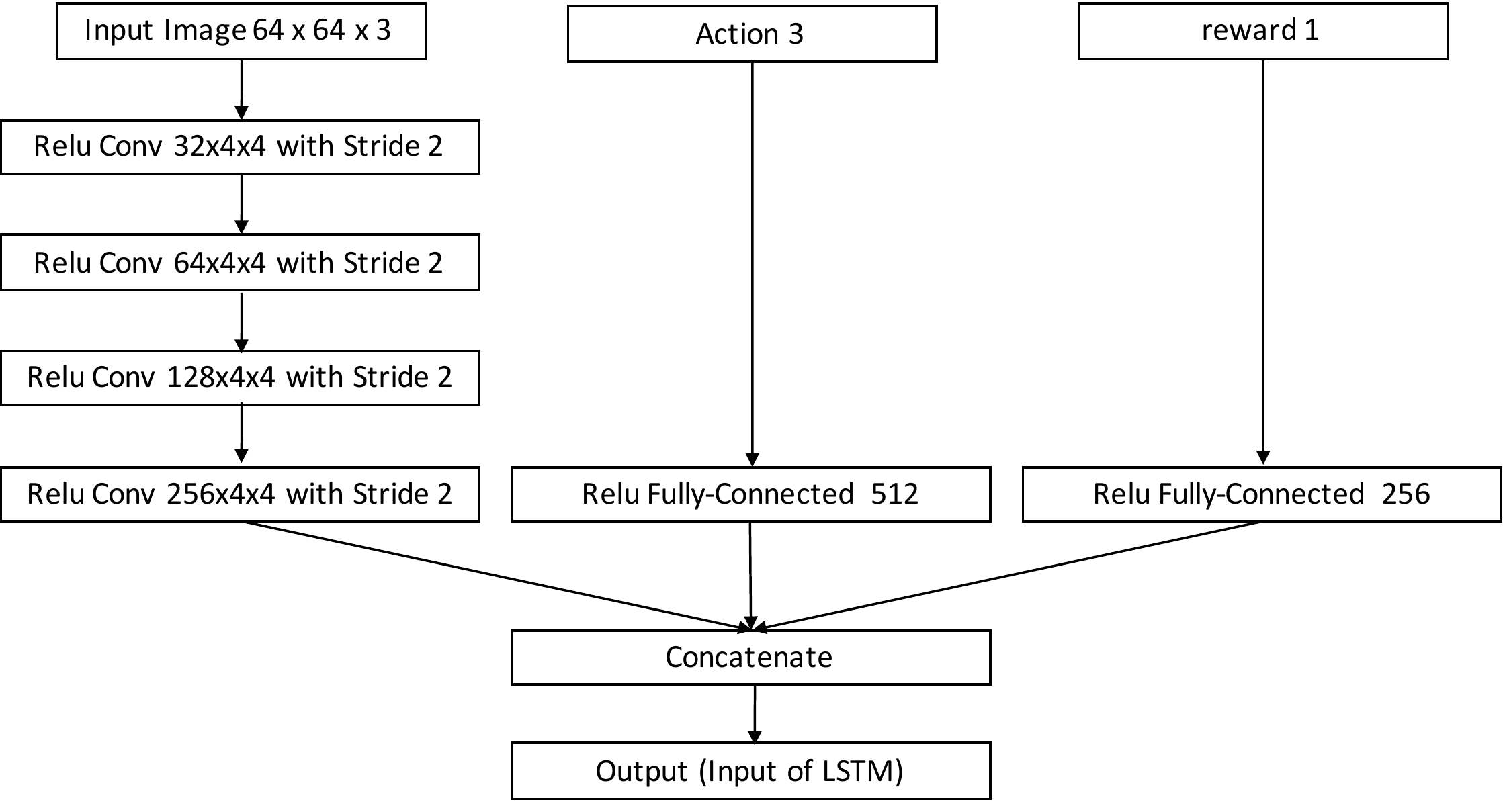} 
    \caption{Network architecture of preprocessor.}
    \label{fig:encoder}
\end{figure}

\begin{figure}[htp]
\centering
\begin{minipage}{.48\textwidth}
  \centering
  \includegraphics[width=.6\linewidth]{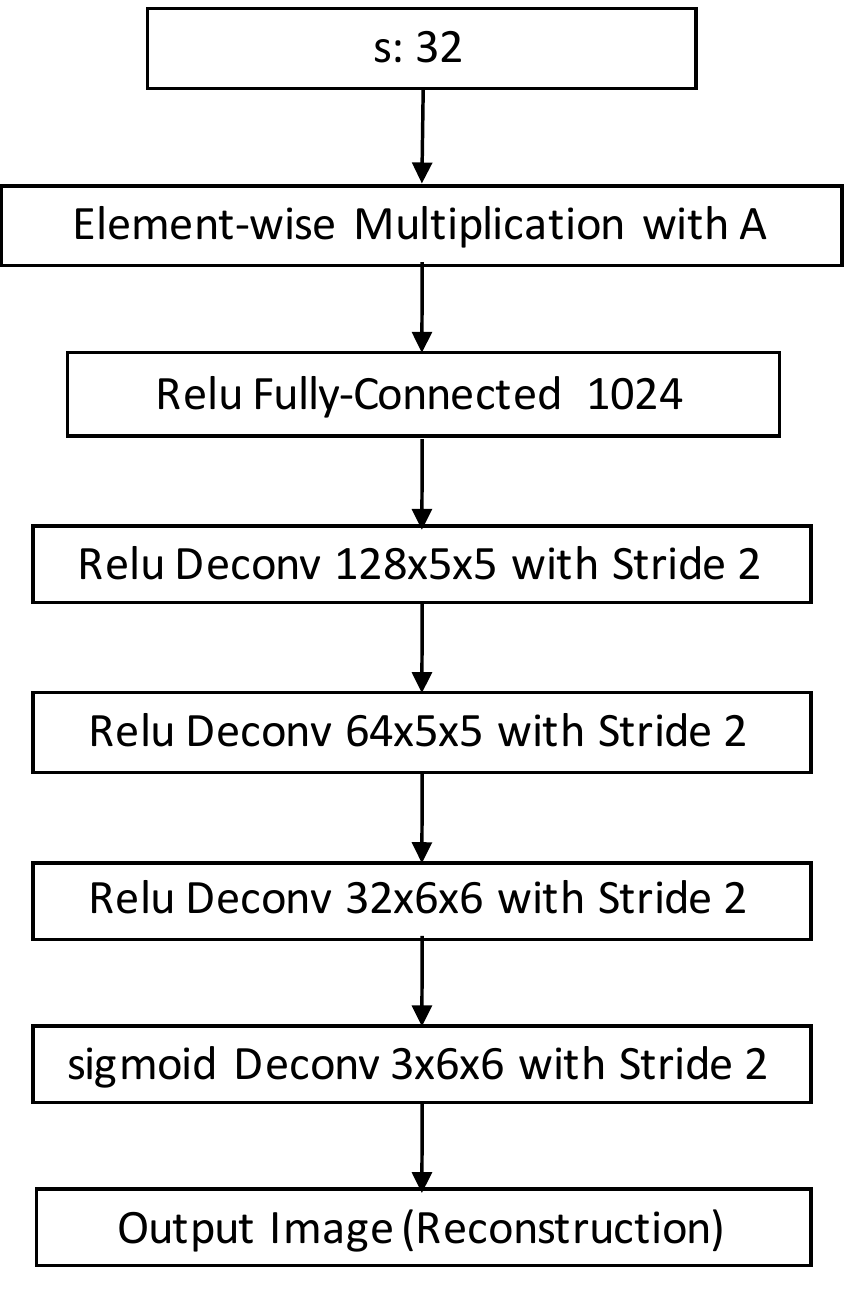}
  \captionof{figure}{Network architecture of observation reconstruction.}
  \label{fig:obs_reconst}
\end{minipage}%
\quad
\begin{minipage}{.48\textwidth}
  \centering
  \includegraphics[width=.8\linewidth]{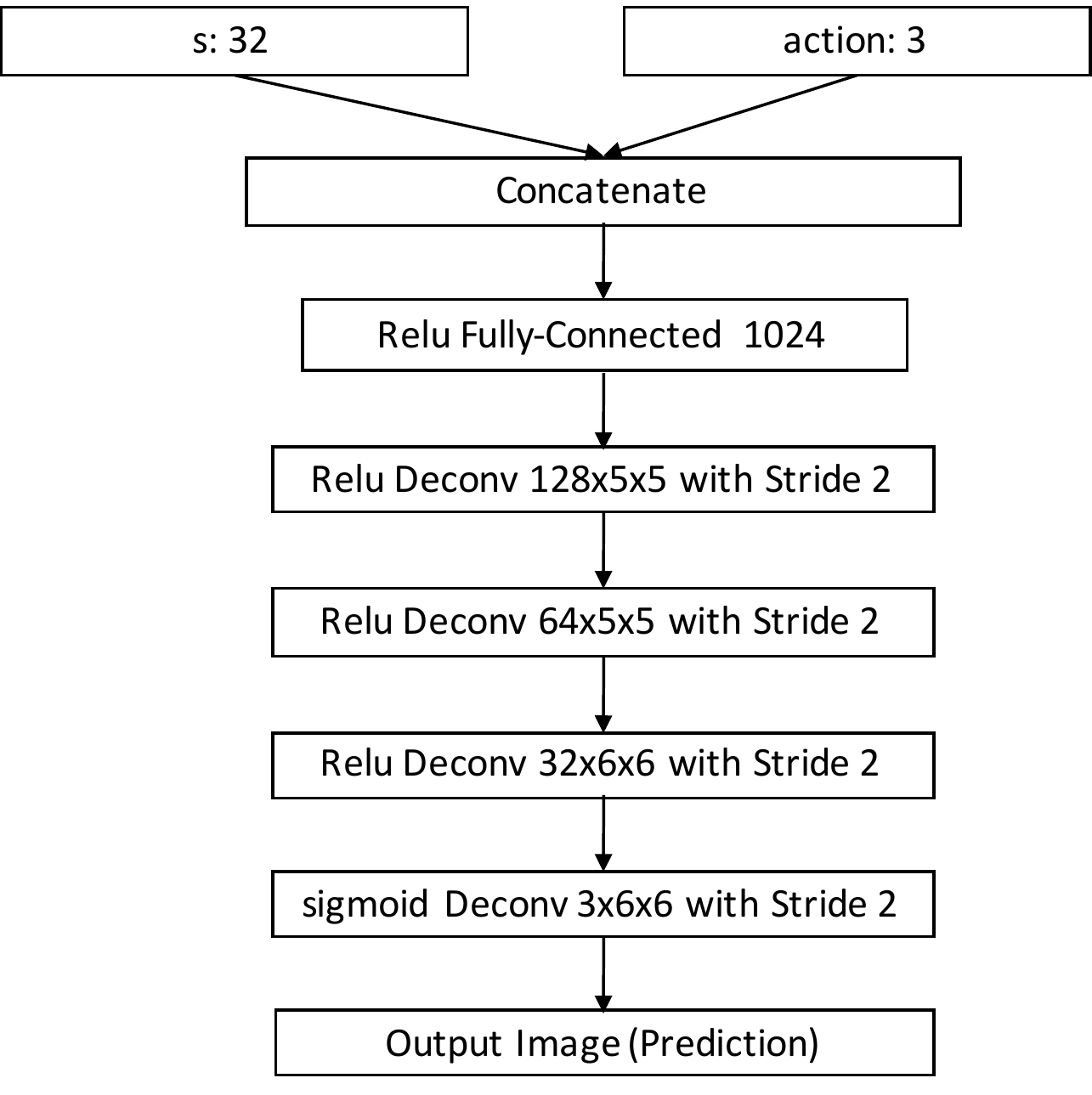}
  \captionof{figure}{Network architecture of observation prediction.}
  \label{fig:obs_pred}
\end{minipage}
\end{figure}

\begin{figure}[htp]
\centering
\begin{minipage}{.5\textwidth}
  \centering
  \includegraphics[width=.9\linewidth]{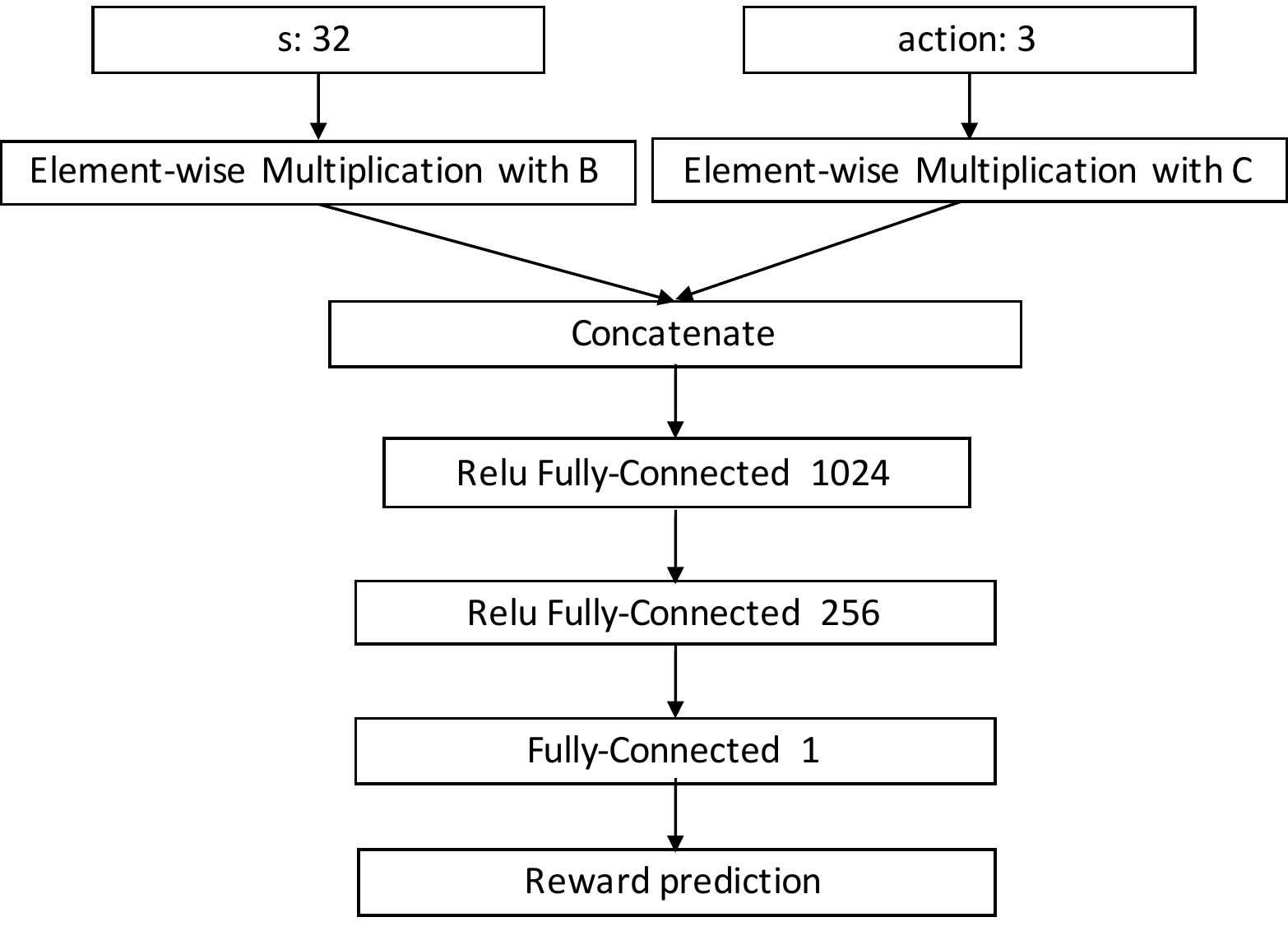}
  \captionof{figure}{Network architecture of reward.}
  \label{fig:reward_pred}
\end{minipage}%
\begin{minipage}{.5\textwidth}
  \centering
  \includegraphics[width=.9\linewidth]{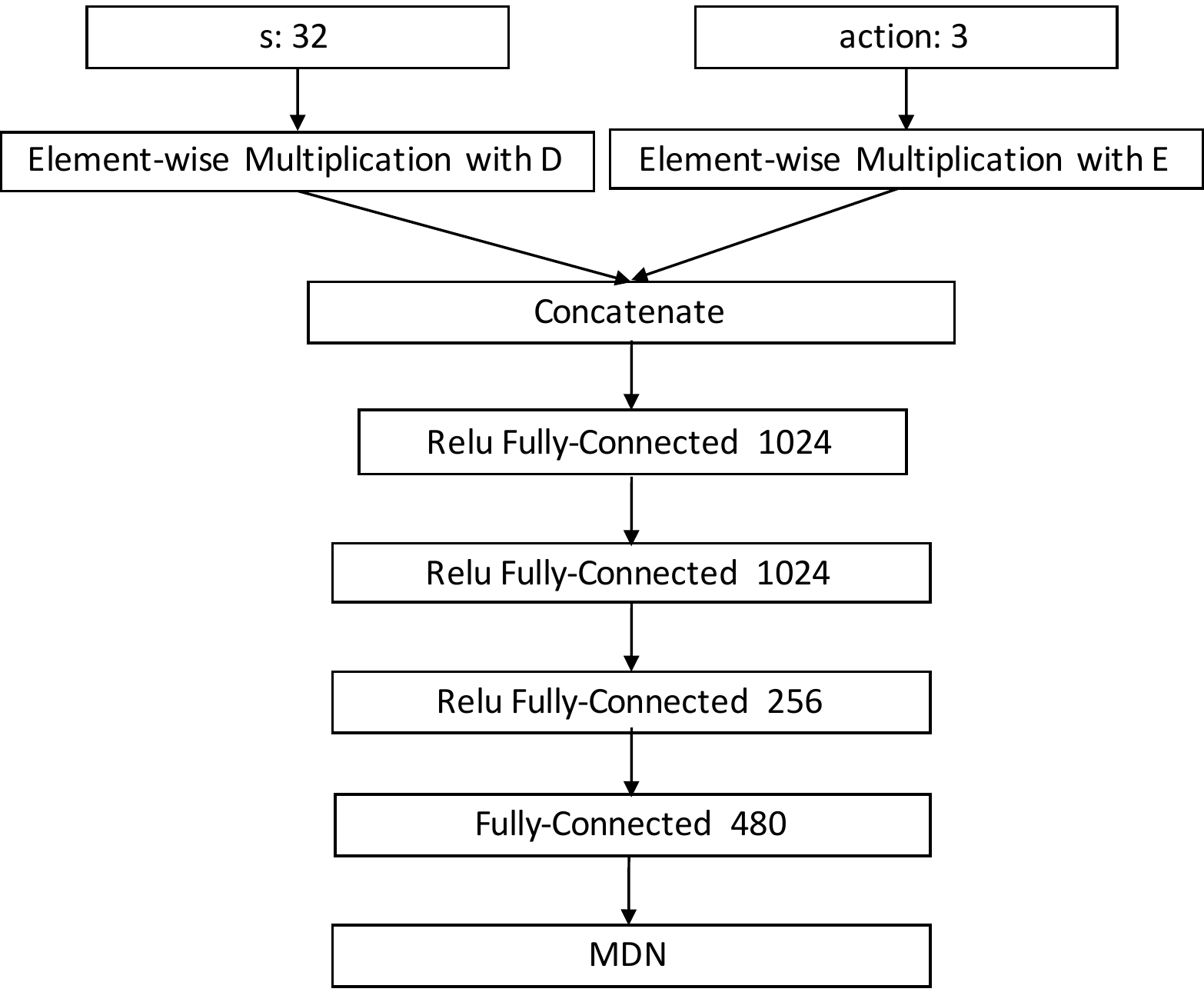}
  \captionof{figure}{Network architecture of transition/dynamics.}
  \label{fig:dynamic}
\end{minipage}
\end{figure}

In the car racing experiment, the original screen images were resized to $64 \times 64 \times 3$ pixels. The encoder consists of three components: a preprocessor, an LSTM, and an MDN. The preprocessor architecture is presented in Figure \ref{fig:encoder}, which takes as input the images, actions and rewards, and its output acts as the input to LSTM. We used 256 hidden units in the LSTM and used a five-component Gaussian mixture in the MDN. The decoder also consists of three components: a current observation reconstructor (Figure \ref{fig:obs_reconst}), a next observation predictor (Figure \ref{fig:obs_pred}), and a reward predictor (Figure \ref{fig:reward_pred}). The architecture of the transition/dynamics is shown in Figure \ref{fig:dynamic}, and its output is also modelled by an MDN with a five-component Gaussian mixture. 
% The architecture of the action prediction is given in Figure \ref{fig:action_pred}, which is a two-layer MLP taking states and rewards as input and predicted action as output. 
In the VizDoom experiment, we used the same image size and the same architectures except that the LSTM has 512 hidden units and the action has one dimension. It is worth emphasising that we applied weight normalization to all the parameters of the architectures above except for the structural matrices $D_{(\cdot)}$.

In DDPG, both actor network and critic network are modelled by two fully connected layers of size 300 with ReLU and batch normalisation. Similarly, in DQN \citep{mnih2013playing} on both ASRs and SSSs, the Q network is also modelled by two fully connected layers of size 300 with ReLU and batch normalisation. However, in DQN on observations, it is modelled by three convolutional layers (i.e., relu conv $32 \times 8 \times 8$ $\longrightarrow$ relu conv $64 \times 4 \times 4$ $\longrightarrow$ relu conv $64 \times 3 \times 3$) followed by two additional fully connected layers of size 64. In DRQN \citep{hausknecht2015deep} on observations, we used the same architecture as in DQN on observations but padded an extra LSTM layer with 256 hidden units as the final layer.

%In each experiment of world models, the controller is parameterized by a fully connected layer whose size is the dimensionality of latent states $\vec{s}_t^{\,\text{ASR}}$. In this case, it has much fewer parameters than the original controller in the world models. 

